\pdfoutput=1
\documentclass[acmsmall,authorversion,nonacm]{acmart}

\usepackage{latexsym}
\usepackage{amsmath,amsfonts}%
\usepackage{amsthm}%
\usepackage{mathrsfs}%
\usepackage[title]{appendix}%
\usepackage{xcolor}%
\usepackage{textcomp}%
\usepackage{manyfoot}%
\usepackage{booktabs}%
\usepackage{algorithm}%
\usepackage{algorithmicx}%
\usepackage{algpseudocode}%
\usepackage{listings}%
\usepackage{graphicx}%
\usepackage{multirow}%
\usepackage{caption}
\usepackage{subcaption}
\usepackage{mathtools}
\usepackage{xspace}
\newcommand{\chistera}{\textsc{Chist-Era}\xspace}
\makeatletter
\def\@copyrightspace{\relax}
\makeatother
\begin{document}

\title{Explainable Active Learning for Preference Elicitation}

\author{Furkan Cantürk}
\authornote{This study was done during his Graduate Research Assistantship at Özye\u gin University.}
\email{furkan.canturk@ozu.edu.tr}
\orcid{0000-0003-4937-6538}
\affiliation{%
  \institution{\"{O}zye\u{g}in University}
   \streetaddress{Ni\i{s}antepe Mah. Orman Sok. No:34-36, Alemda\u{g}, \c{C}ekmek\"{o}y}
  \city{Istanbul}
  \country{T\"{u}rkiye}
  \postcode{34789}
}

\author{Reyhan Aydo\u{g}an}
\orcid{https://orcid.org/0000-0002-5260-9999}
\affiliation{%
  \institution{Özye\u gin University}
  \streetaddress{Ni\i{s}antepe Mah. Orman Sok. No:34-36, Alemda\u{g}, \c{C}ekmek\"{o}y}
  \city{Istanbul}
  \postcode{34794}
  \country{T\"{u}rkiye}}
\affiliation{
  \institution{Delft University of Technology}
  \streetaddress{Van Mourikbroekmanweg 6}
  \city{Delft}
  \state{}
  \postcode{2628XE}
  \country{The Netherlands}}
\email{reyhan.aydogan@ozyegin.edu.tr}

\renewcommand{\shortauthors}{Cantürk and Aydo\u{g}an}

\begin{abstract}
Gaining insights into the preferences of new users and subsequently personalizing recommendations necessitate managing user interactions intelligently, namely, posing pertinent questions to elicit valuable information effectively. In this study, our focus is on a specific scenario of the cold-start problem, where the recommendation system lacks adequate user presence or access to other users' data is restricted, obstructing employing user profiling methods utilizing existing data in the system. We employ Active Learning (AL) to solve the addressed problem with the objective of maximizing information acquisition with minimal user effort. AL operates for selecting informative data from a large unlabeled set to inquire an oracle to label them and eventually updating a machine learning (ML) model. We operate AL in an integrated process of unsupervised, semi-supervised, and supervised ML within an explanatory preference elicitation process. It harvests user feedback (given for the system's explanations on the presented items) over informative samples to update an underlying ML model estimating user preferences. The designed user interaction facilitates personalizing the system by incorporating user feedback into the ML model and also enhances user trust by refining the system's explanations on recommendations. We implement the proposed preference elicitation methodology for food recommendation. We conducted human experiments to assess its efficacy in the short term and also experimented with several AL strategies over synthetic user profiles that we created for two food datasets, aiming for long-term performance analysis. The experimental results demonstrate the efficiency of the proposed preference elicitation with limited user-labeled data while also enhancing user trust through accurate explanations.
\end{abstract}

\begin{CCSXML}
<ccs2012>
   <concept>
       <concept_id>10002951.10003317.10003331.10003271</concept_id>
       <concept_desc>Information systems~Personalization</concept_desc>
       <concept_significance>500</concept_significance>
       </concept>
       <concept>
       <concept_id>10010147.10010257.10010282.10011304</concept_id>
       <concept_desc>Computing methodologies~Active learning settings</concept_desc>
       <concept_significance>300</concept_significance>
       </concept>
 </ccs2012>
\end{CCSXML}

\ccsdesc[500]{Information systems~Personalization}
\ccsdesc[300]{Computing methodologies~Active learning settings}

\keywords{Active Learning, Explainable AI, Preference Elicitation, Interactive Recommendation}

\maketitle

\section{Introduction}

Modeling user preferences plays a crucial role in interactive intelligent systems such as agent-based negotiation systems \cite{aydogan_learning_2012} and especially in recommendation systems (RSs) \cite{bobadilla2013recommender}. The effectiveness of user preference elicitation and modelling approaches becomes evident in cases of the cold-start problem in recommender systems \cite{Lika2014,Ricci2015}, where the system does not have sufficient information to generate personalized/well-targeted recommendations when new users arrive. 
Nevertheless, solutions to this problem often rely on the availability of data from other users in the system, such as reviews, clicks, purchases, and ratings.

In this study, we address the cold-start problem of recommendation systems, particularly where the system does not have any or sufficient number of users yet or is not allowed to access existing users' data (e.g., health information of individuals might not be agglomerated in for assistant systems). Therefore, it is necessary to carry out a preference elicitation step in recommendation systems to acquire information about users' preferences through some form of interaction with users. This step should be designed to maximize information acquisition from the user with a minimal cognitive workload for the user. Moreover, the design should follow some criteria considering user experience since users can be frustrated if the preference elicitation step is a significantly time-consuming process. Labeling a high number of items or features, providing quantitative information for a set of items, ranking a huge number of items at once, and any similar quantitatively or qualitatively dense evaluation process might increase the cognitive workload significantly, which should be avoided in the system design. In this study, our objective is mainly to handle the trade-off between maximizing the information received from the user and minimizing the user's effort to provide that information. 

Active Learning (AL) is a branch of machine learning (ML) in which informative samples among large unlabeled data are selected to ask an oracle (e.g., an end-user or a domain expert) to label or annotate them \cite{Tong2001}. Subsequently, an ML model is updated by including those informative samples into the training dataset. As getting information from users to learn their preferences can be tiresome for them, AL could be a convenient information acquisition method for preference elicitation \cite{Rubens2015,Iovine2022}, considering the cost of obtaining labels for items from users.
In addition to AL, we consider the explainability of RS as a second component to fertilize the information acquired through user interactions. Conversational or interactive RSs can ask users to verify whether the system's explanations regarding the reasoning of the estimated preference model with ML are consistent with their preferences. Their feedback could be utilized to personalize the preference model over time. In this way, users can play an active role in tailoring recommendations to their interests and preferences by responding to the explanations through interactive RSs. Rather than relying on only user labels or ratings for items, the preference model can be explicitly adapted to user preferences, thus, amended faster as much as utilizing the user feedback.

Drawing upon the advantages of AL and explainable AI, this study assesses the potential of Explainable Active Learning (XAL) for eliciting user preferences in RSs. Though XAL is originally proposed for machine teaching \cite{Liao2021}, it presents a promising avenue to efficiently personalize RSs by asking users to provide feedback on informative instances in response to the system's explanations. Accordingly, we propose a preference elicitation framework tailored for the learning task considering these two components: AL and explainability. At the start-up of the system, the user experiences a preference elicitation phase in which she is inquired to label `informative instances' chosen by the system so that a minimal amount of user-labeled items can be enough to leverage an underlying preference learning model. The latter feature of our framework is providing a preference-based explanation associated with the presented/recommended item. The user is inquired to give feedback on explained recommendations, and this feedback is used to update the preference model. In this way, explanations of the recommendations also indirectly serve for personalizing the system and building a trusted experience in the system meanwhile.

We measure the effectiveness of the proposed solution to user preference elicitation for food RSs by conducting human experiments and long-term empirical analysis of several AL strategies with synthetic user profiles that we generated for two food datasets. 
Hereby, our study contributes to handling the cold start problem in RSs in several aspects. The proposed XAL-based methodology enables preference elicitation without depending on the availability of any type of historical data in the system. Secondly, it significantly accelerates the learning of user preferences by (i) using user feedback to update the underlying ML model and (ii) requiring much fewer user interactions (in other words, fewer data labeling) compared to the baseline (random) sample selection. Moreover, the explainability of the system serves for personalization by incorporating user feedback into the estimated preference model with ML. The human experiments validate that the proposed preference-based explanation schema is comprehensible for humans, and the cognitive workload of interacting with the system is tolerable. Last but not least, we implement five AL strategies and benchmark them to reveal how they perform through the synthetic profile experiment. The source code of the proposed framework is accessible online%
\footnote{\url{https://github.com/furkancanturk/explainable_active_learning}}.

The rest of the paper is organized as follows. Firstly, Section \ref{sec-related-work} reviews the related work on preference elicitation and AL in RSs. Section \ref{background} gives the necessary background about XAL and also highlights the cost of user interaction. We present our proposed framework in Section \ref{sec-framework} and evaluate it in the food recommendation domain with human experiments and empirical analysis. Finally, Section \ref{sec-conclusion} concludes the study with a discussion involving future work directions. 
\section{Related Work}
\label{sec-related-work}

%%%
% Pecune2020: A Recommender System for Healthy and Personalized Recipes Recommendations
% Gao2021: Advances and Challenges in Conversational Recommender Systems: A Survey
% Liu2019: Review of Intent Detection Methods in the Human-Machine Dialogue System
% Kadziski2021: Active learning strategies for interactive elicitation of assignment examples for threshold-based multiple criteria sorting
% Teso2019:
% Phillips2018

RSs can exploit traces of user activities on the system, such as query history, click history, reviews, ratings, and purchases. The absence of such data (typically by newcomers) is called the cold-start problem in RSs. A certain amount of user data allows collaborative filtering algorithms to exploit the similarity of users' preferences, content-based filtering algorithms that rely on the similarity of items, and hybrid approaches taking advantage of both filtering. For domains where user activity on the system is relatively low, conversational RSs represent an alternative solution to the cold-start problem \cite{Gao2021}. It utilizes prior knowledge of the recommendation domain, often depending on relations between entities, to construct an ontology/knowledge base \cite{Haussmann2019}. Such ontologies are dedicated to providing results that satisfy users' queries.

In the case of literally no user data, an initial preference elicitation step is carried out with interactive recommendation approaches. These approaches can apply several channels for the acquisition of user knowledge, such as demographic filtering \cite{Adomavicius2005}, presenting a set of features or contexts to be selected \cite{Druck2009, Khan2021}, providing a set of example recommendations to be labeled \cite{Pecune2020}, and question-based preference elicitation \cite{Kostric2021,Buzcu2022} or intent detection via intelligent dialogue systems \cite{Liu2019}. An initial interaction phase with the user supply data that enable the execution of subsequent recommendation models. Rather than relying on a static latent or explicit variable matrix representing the user profile, RSs can be updated continuously by interacting with the user, which enables a personalized experience with the system. Especially advancements in Natural Language Programming and knowledge graph-based learning convert static RSs into an interactive form with users and facilitate learning a broad view of preference aspects \cite{Zhang2020}. So, not only meeting users' pleasure but also meeting their true needs and detecting intentions or specific goals of them can be possible by such interactive intelligent systems \cite{Liu2019}. 

Our study focuses on the design of interactive preference elicitation in RSs. Interactions with users should maximize information acquisition with minimal workload to overcome user frustration. This objective is in line with AL, which is often employed in ML tasks when dealing with scenarios involving limited labeled data or extensive unlabeled data streams. In the scope of AL, informative instance selection methods \cite{Fu2012} construct a limited training sample that maximizes performance ML models. Therefore, our study envisions that AL can serve to handle the new user problem in RSs. 

Rubens et al. describe existing AL methods with the viewpoint of the classical new-user problem in RSs that considers ratings by available users to estimate a rating vector of a user over all items \cite{Rubens2015}. Such studies on AL for RSs rely on data of available users in the systems, following the idea of collaborative filtering \cite{Elahi2014,Elahi2016,Iovine2022}. For example, rating variance, popularity, rating entropy, and log-popularity-entropy, which balances popularity and entropy, are the metrics to determine which instances are selected to ask the new user to label them. These metrics are calculated based on the ratings and/or activity of other users in the system. Alternatively, Iovine et al. also use AL with content-based filtering by selecting dissimilar items from pre-existing items in the user profile for rating them by the user \cite{Iovine2022}.
Despite AL being relatively more adopted with collaborative filtering, Hernández-Rubio et al. propose a hybrid filtering approach with AL, Aspect-based AL \cite{HernndezRubio2020}. It uses collaborative information and \textit{aspect similarity} among items together where item aspects are extracted from user reviews. The study benchmarks the proposed strategy with common AL strategies for collaborative filtering. A similar approach proposed by Bu and Small exploits relations among items and also uses Fisher Information-based AL \cite{Bu2018}. 
While all these AL approaches rely on available explicit or implicit information (feedback) about user preferences, our study tackles the issue of the complete absence of any user data within the system, which those approaches cannot be referred to solve. 
The principal function of the proposed framework in our study is to conduct AL without any labeled data at hand and without relying on data from other users. The foundation for determining the informativeness of recommended items with respect to user preferences is based on unsupervised, semi-supervised, and supervised learning models, whereas prior AL approaches rely on statistical information derived from the presence of user data. Besides, our XAL-based preference elicitation approach explains how the supervised ML model estimates user preferences, aiming to refine the model by obtaining user feedback.

%unsupervised, semi-supervised, and supervised learning
Apart from collaborative filtering accompanied by AL for RSs, content-based filtering is used with AL for information retrieval in \cite{Zhang2002}. In this study, an annotator is inquired to check whether an instance has a given attribute to serve the task of finding the best-matching instances in a database according to sought attributes. 
This approach is similar to AL with labeling relevance of item feature \cite{Druck2009}, which serves for identification of effective features for user preferences.

When considering how to formulate user feedback explaining preferences in XAL for RSs, a usual way could involve rating recommendation items. However, user ratings might not be reliable information for preference elicitation \cite{Gajos2009, Kalloori2018} as providing consistent and sufficiently differentiating ratings for all items requires a complex cognitive process. For example, the user can rate an item with the highest score and can also rate another item that s/he likes more than the previous one with the highest score again. This issue is not only about the noise in the user ratings. Humans could not easily and always make a consistent comparison within a large set of items. Instead, pairwise item evaluation can provide more reliable preference information \cite{Jones2011}. 
Another basic user feedback process is the identification of relevant or salient features of items for user preferences. In \cite{Teso2019}, when a prediction by an ML model is correct but with incorrect reasoning (explanation), the model is retrained with training data added with counterexamples aligned with the correct reasoning provided by an annotator. However, in this framework, the annotator verifies only the relevance of features for the prediction, not providing how the features affect. Gajos et al. examine whether asking relevance of features for user preferences is the most efficient way to receive user feedback. They conclude that some other structured feedback forms provide modestly but consistently improved predictive accuracy in learning user preferences compared to identifying the feature relevance \cite{Gajos2009}.
In addition to the complexity of user feedback (or preference-based explanation), the incorporation of knowledge distilled from user feedback into a recommender model represents the subsequent issue in interactive learning for preference elicitation. In our study, explanations for the system's recommendations are structured as labels for item features, and the user is expected to provide feedback on the explanations by correcting the feature labels according to her preference. 

In our context, as user feedback will be utilized for updating the underlying ML model for estimating user preferences, the structure of user feedback should permit the incorporation of preference information into ML models as well as being easy to construct for users. Therefore, we propose representing user feedback as a categorization of attributes in the recommendation domain based on their effect on user preferences, such as liking yogurt and disliking spicy sauce in foods.
%To learn users' preferences better, they can be inquired to provide explicit feedback for the recommendations by the system. 
Alternatively, explicit user feedback can be structured as a set of rules or liked/disliked item features. 
%In fact, the form of system explanations also should be aligned with the user feedback type. 
Incorporating user feedback into a recommendation model allows users to personalize recommendations on their own; thus, it establishes a more satisfying system experience. Human feedback is incremental, incomplete, imprecise, and heterogeneous \cite{Liao2021}; therefore, evaluation of preference explanations by the system based on lots of features or quantitative information could be exhausting for users.

\section{Background}
\label{background}

This section first provides preliminary information about AL, then discusses different forms of preference explanation considering the cognitive workload of providing feedback to the system and the benefit of information level (i.e., local or global) for the preference elicitation process. 

\subsection{Explainable Active Learning}

AL is an ML technique that involves an oracle labeling given samples for the training models and a \textit{smart selection "strategy} determining which samples are to be labeled. It is devised to deal with cases of abundant unlabeled data or limited labeled data at hand, costly data labeling, and evolving stream data, which are typical problems faced in real-world ML applications. AL consists of three stages: \textit{(i)} informative instance selection, \textit{(ii)} labeling/annotation, and \textit{(iii)} model update. Firstly, a query (sampling) strategy selects a set of samples about which the predictor model is the least experienced/certain. In the second step, an oracle (generally a human annotator) is asked to label these samples. Finally, the predictor model is updated based on the oracle's input. 

There are two fundamental approaches to AL sampling strategies: \textsl{uncertainty-based} and \textsl{diversity-based} sampling \cite{Monarch2021}. The uncertainty-based strategy aims to strengthen the model's certainty over data near its decision boundaries by supplying informative samples. The question of how to determine such samples or define model uncertainty can be addressed by various uncertainty metrics. The well-known uncertainty metrics in AL based on posterior probability by a predictive model $\theta$ for a given sample $x$ can be listed as the least confidence score, margin score, and entropy \cite{Monarch2021}. Equations \ref{least-confidence}, \ref{margin-score}, and \ref{entropy} define those metrics respectively, where $y_1$ is the most probable class label, $y_2$ is the second most probable class label, and $C$ is the set of class labels. These three metrics are normalized to be in the range of $[0,1]$, where the highest uncertainty score is one. For example, if a binary classifier yields 0.5 posterior probability of being positive (or negative) for $x$, these metrics would yield one, denoting the maximum uncertainty level. 

\begin{gather}
    \text{Normalized Least Confidence} : \frac{|C|}{|C|-1}(1 - P_{\theta}(y_1 \vert x)) \label{least-confidence} \\
    \text{Margin Score} : 1 - (P_{\theta}(y_1 \vert x) - P_{\theta}(y_2 \vert x)) \label{margin-score} \\
    \text{Normalized Entropy} : \frac{- \sum_{i \in C} P_{\theta}(y_i \vert x)log P_{\theta}(y_i \vert x)}{log \vert C \vert} \label{entropy}
\end{gather}

The lack of the model to find the decision boundaries correctly can be sourced by \textsl{aleatoric} and \textsl{epistemic} uncertainty. Aleatoric uncertainty arises from inherent noise in the model output itself (i.e., the predictive error by model output variance), while epistemic uncertainty is due to the lack of information --- knowledge gap between in-sample and out-of-sample data. Aleatoric uncertainty of a model can be alleviated by refining its decision boundaries, i.e., retraining the model with a support set augmented with a labeled sample around the boundaries. However, when the model faces instances outside of the training data distribution, it extrapolates to explain them, which results in epistemic uncertainty. Epistemic error is the problem addressed by ML applications in constantly changing environments where the distribution of new data shifts. To reduce the knowledge gap of a model, samples providing different information are needed to feed the model. Hence, \textsl{diversity-based sampling} is used as another sampling strategy for AL that selects samples from different regions of data. Diversity-based sampling aims to select samples that are representative and diverse in nature. This approach prevents the model from becoming biased towards a specific subset of the data and ensures that it gains a comprehensive understanding of the data distribution. To ensure the diversity of samples to be asked to an oracle, unsupervised learning is a natural candidate method to capture general patterns in data without labels. Diversity-based approaches can be generalized under \textsl{cluster-based sampling} and \textsl{representative sampling}. Cluster-based sampling facilitates the acquisition of information about the global structure of data, while representative sampling cares about differences between labeled data in the training set and unlabeled (target) data. To ensure information acquisition over both representative and uncertain samples, uncertainty and diversity-based sampling approaches can be combined in various ways, which are extensively described in \cite{Monarch2021}. 

Each AL strategy described so far basically tries to maximize model performance over entire unlabeled data in different ways. In fact, a generic sampling strategy is designed based on \textsl{Expected Error Reduction} \cite{Settles2009}, which is the true objective of all AL strategies. Those strategies select samples to minimize the error that occurs by predictions for a validation set as the expected error for unlabeled data under the assumption of no distribution shift between labeled and unlabeled data. To overcome re-training cost by the expected error reduction approach, which limits its practicality, statistical techniques like \textsl{Monte Carlo Estimation} and \textsl{Influence Functions} are used to estimate the error \cite{Roy2001, Teso2021}. Error reduction-based AL strategies necessitate a representative validation set. However, labeled data at hand can be scarce, and thus creating a sufficiently representative validation set from available labeled data could not be possible, which makes the strategy inapplicable for earlier phases of AL. In other words, data maturity is a prerequisite for implementing error-reduction-based strategies. 
Since our study addresses the problem of preference elicitation in the case of no data beforehand, labeled data obtained after initial interactions with the user cannot be mature enough to feed supervised learning (SL) models. The scarce labeled data problem is also addressed by semi-supervised learning (SSL). SSL is an ML technique that utilizes a distance metric referring nearby instances likely to have the same labels and disseminates available label information toward close (similar) instances. Therefore, SSL is a well-suited component for AL by augmenting training data for SL models with pseudo-labeled data \cite{Chen2017,Zhu2003,Gu2014,Gao2020}.

Fusing explainable AI (or interpretable ML) and AL techniques, XAL is an ML concept recently proposed for machine teaching \cite{Liao2021}, which aims to learn tasks efficiently and autonomously with minimal human intervention. By leveraging human expertise and interaction, the hope is to accelerate the training process, reduce the amount of required data, and improve the overall performance of ML algorithms. 
A learning concept similar to XAL is \textsl{Coactive Learning} (CL) \cite{Shivaswamy2015}, where an oracle responds by providing an improved configuration of system prediction results. CL aims to minimize the regret (i.e., the utility difference between the system prediction and respective improved configuration by the user) throughout the interaction steps. An implementation of the idea of XAL for preference elicitation is done for trip planning by \textsl{Coactive Critiquing} \cite{Teso2017}. Coactive Critiquing is a constructive CL approach that discovers feature space iteratively, obtaining features from the user. In each interaction, the user provides a better item configuration in place of the recommended one by modifying it or introducing new features. Consequently, a linear utility function is learned over the discovered features throughout interactions of CC. The utility function is updated based on whether it performs pairwise comparisons correctly over the set of item pairs (i.e., the pairs of a recommended item and the corresponding improved item configuration by the user). 

\subsection{Local versus Global Explanations on Preferences}

Preference elicitation in recommendation systems is the process of retrieving information about users' preferences, likes, and dislikes to create personalized recommendations. This involves obtaining explicit feedback like ratings or implicit feedback from user behaviors such as clicks and purchases. It addresses challenges like the cold-start problem for new users or changing preferences over time, enhancing recommendation accuracy. During this process, the users are asked to express what they prefer and to what extent they prefer. Besides, they can explain why they prefer those alternatives/choices. 

In this study, we formalize preference-based explanations for recommendation systems with the principle that users provide feedback in the same form as the explanations provided by the system. This principle makes the feedback process harder for users as the complexity of the explanations increases. Therefore, how the system's explanations for recommendations are structured is a factor in the cost of information retrieval through user interactions. Accordingly, we focus on eliciting preferences by means of expressing general judgment in the given recommendation context, such as ``I like foods with tomatoes.'', `` I don't like spicy foods.'', and "Fat content in food is unimportant to me." 
Without a doubt, someone also prefers expressing their preferences with a granular comparison of attributes within a specific item, such as ``I prefer chicken over tuna in sandwiches.'', ``I like having mushrooms on the piazza more than peppers.''.
These explanations offer information at a local level, meaning that each item feature is assessed in relation to others. To input such fine-grained `local information' to the system, users can rank or assign scores for features of each presented item separately. However, expressing preferences in that way could be frustrating for users since each presented item requires a comparing effect of features relative to each other on their preferences.  
On the other hand, expressing general preferences is a more straightforward process, thus requiring relatively less cognitive workload for the users.
Besides, getting feedback from users at a global level could foster preference elicitation more than using local information since the underlying ML model for learning user preferences can be accurate as much as it is supplied with information valid for the entire decision space (or recommendation domain). After reaching a mature predictivity, the estimated preference model could be refined on local reasoning.

Since our study aims to elicit the preferences of new users, in principle, their general preferences should be learned by the system so that the underlying ML model estimates their preference model accurately as much as possible with a few labeled data by the users.
Therefore, we design the explainability of the proposed framework based on prioritizing the refinement of global reasoning of ML models by inputting user feedback. Similarly, the explainability of ML models can be provided at a global or local level.
That is, a \textit{local explanation} presents information about how a model makes a prediction for a specific instance. It aims to provide insights into why the model made a particular decision for that particular input.  For example, a local explanation formed of negative and positive weights for features of a given item presents how the model evaluates those features to estimate the likeability of that item by the user.
In contrast, a \textit{global explanation} describes the general behavior of the model over the entire domain. It helps users understand the broader patterns and relationships that the model has learned from the data. Global explanations are useful for identifying general trends, biases, and important features that consistently influence the model's predictions. Accordingly, weights assigned to all features available in the domain could be used for understanding how the ML model makes decisions for the given task, preference modeling.
\section{Explainable Active Learning-based Preference Elicitation Framework}
\label{sec-framework}

Before the recommendation phase, eliciting and modeling new users' preferences (i.e., what they like and why) could contribute to dealing with the cold-start problem in recommendation systems. The aim is to find a function mapping a set of items $\mathbf{X}$ to labels $\mathbf{y}$ where each $\mathbf{y}_i$ could be a rank or a binary value denoting whether or not the user likes item $\mathbf{x}_i$. 
Each item $\mathbf{x}_i$ in $\mathbf{X}$ is represented as a binary vector $<d_0, d_1, \ldots, d_n>$, indicating whether the item contains features $d \in D$ from the recommendation domain, where $D$ is the set of all available features in the domain and $n = ||D||$. In other words,  $\mathbf{x}_i$ is a feature vector having ones for the contained features $D_i$ in item $i$ and zeros for the not-contained features.
In such a setting, it can be assumed that an end-user of a recommendation system, $\mathcal{O}$, can label any item in the system $\mathbf{x}$ and explain how the features of items affect their preference model. However, this assumption is unrealistic due to the complexity of the process (i.e., asking users to label thousands of items and explain how each feature affects their preferences is not feasible). Therefore, only a small set of items $L_t \subset X$ at time step $t$ could be labeled by the users. Consequently, an approximate preference function could be modeled by only considering a small set of labeled data, $\hat{f}: L_t \rightarrow \mathbf{y}$. The chosen set significantly influences the preference modeling performance, which should be accurate for the entire domain, $\mathbf{X}$. Hence, we need a practical filtering function $\phi$ selecting samples from $\mathbf{X}$ to be labeled by $\mathcal{O}$. Accordingly, this work proposes an AL approach for $\phi$ relying on interactions with the user and updating an ML model for $\hat{f}$ based on user feedback in an explainable RS.

At the initial state of the system, no user-labeled data is in hand. Therefore, the user is inquired to label a sample of items in the recommendation domain. The diversity of the initial sample is ensured by deploying an AL method based on unsupervised learning (UL) so that user preferences over the entire recommendation domain can be captured as much as possible. However, the size of the initial labeled data by a human effort at once would not be sufficient to model user preferences with an ML model. Therefore, labeled data augmentation is applied with a semi-supervised learning (SSL) model before training a supervised learning (SL) model. SSL produces pseudo-labeled data by exploiting the similarity of instances. An essential assumption about similarity-based learning models is similar (nearby in data space) instances are more likely to be in the same class, which also aligns with human preference/judgment about items (Recall the content-based filtering exploits item-item similarity). After training the model, an interactive step is carried out with the user, which is detailed below.

\begin{figure}[h]
    \centering
    \includegraphics[width=\textwidth]{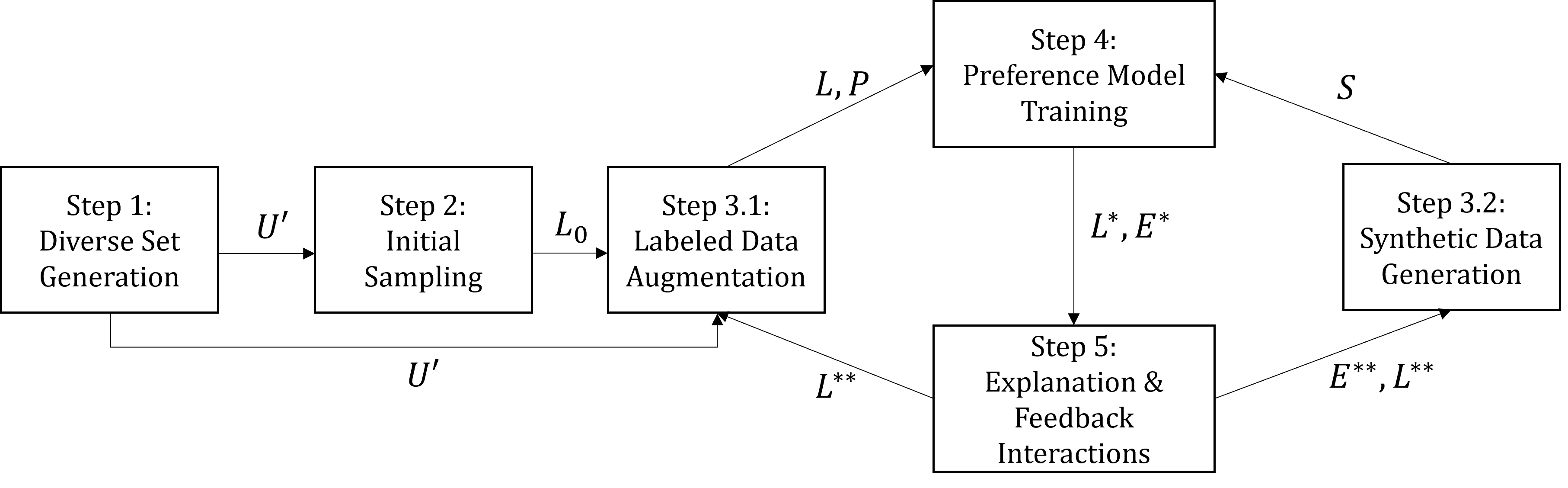}
    \caption{Explainable Active Learning-based Preference Elicitation Framework}
    \label{fig:framework}
\end{figure}

As depicted in Figure \ref{fig:framework}, there are mainly five steps (to be described in detail later) applied as a framework of learning and sampling mechanisms. Steps 1, 2, 3.1, 4, and 5 are implemented sequentially as a preference elicitation phase in the recommendation system, where the user is asked to label selected instances by AL in Step 2. The other interaction is conducted in Step 4, where an initially trained preference model presents explanations and predicted labels of a batch of instances that can be likable or dislikeable to the user in return for getting feedback on the predictions and the explanations.
After completing this phase, this methodology can function to refine the user preference model by applying Steps 3 and 4 after each batch of explanation-feedback interactions on recommendations. 

\begin{figure}[h]
    \centering
    \includegraphics[width=0.4\textwidth]{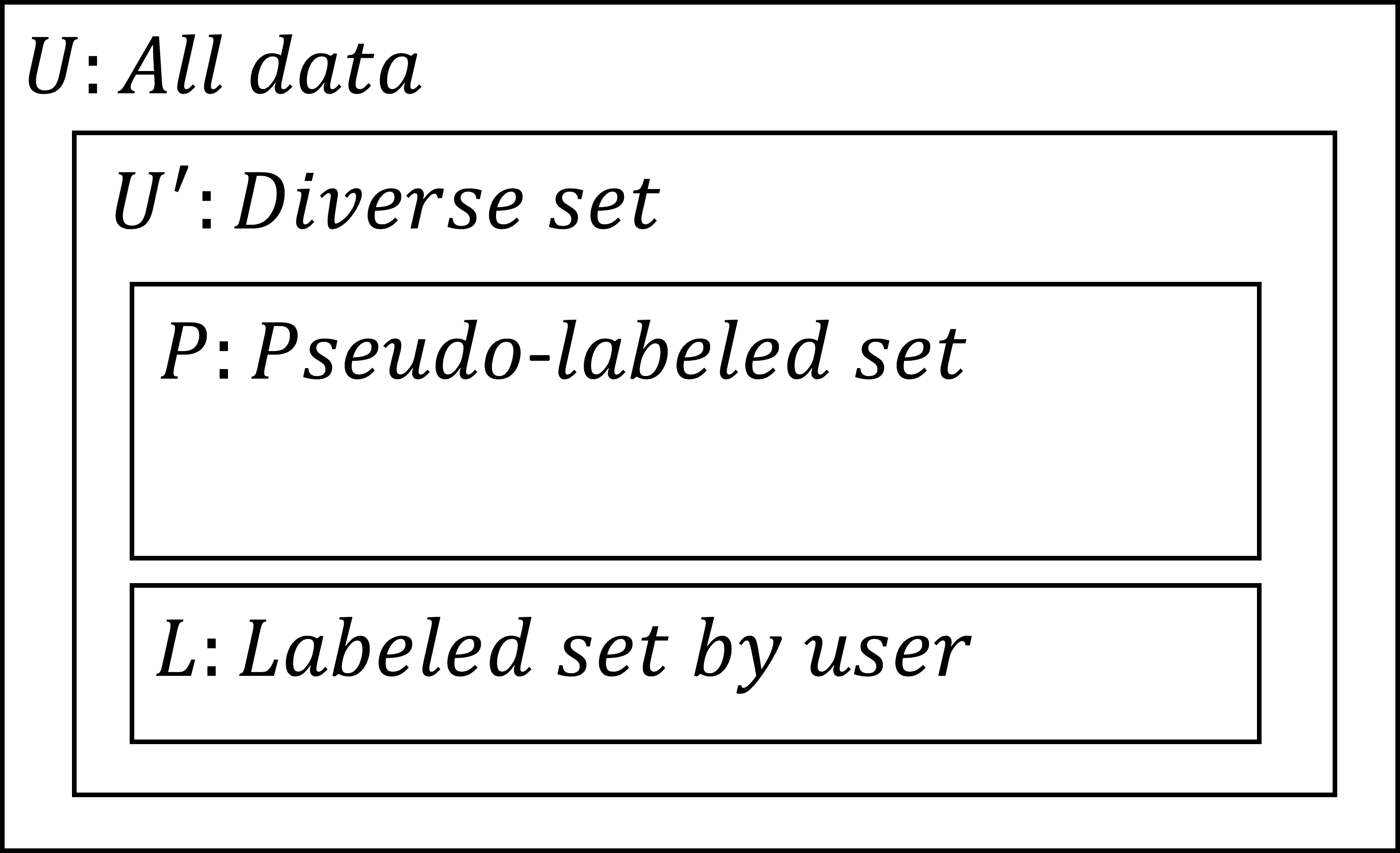}
    \caption{Hierarchy of the Data Sets in the Proposed Framework}
    \label{fig:sets_hierarchy}
\end{figure}

The framework works by partitioning unlabeled data (i.e., of which no user preference information available) into multiple sets whose hierarchical relationship is depicted in Figure \ref{fig:sets_hierarchy}. Here, each set is represented as a separate rectangle annotated with a respective symbol at the top-left corner. We define the presented sets as follows.
An (initially) unlabeled set of items $U$ is inputted into the framework. A diversity sampling method (i.e., a UL-based AL model) produces $U'$ out of $U$ as a candidate set to be labeled by the user and to extract any estimated information about all data, $U$. $U'$ is needed as a representative subset data for large domains because $U$ might be too large (as in RSs) to train learning models and extract statistical information like pairwise item similarities, which is $\mathcal{O}(n^2)$ where $n$ is the number of items. The minimum required size of $U'$ to capture the user's preferences across all items in $U$ is proportional to the size of $U$. As the third set, $L$ is the set of instances labeled by the user and initially $L=\emptyset$. Instances in $U' \setminus L$ are pseudo-labeled via SSL using $L$. The instances labeled by SSL with a confidence level constitute a pseudo-labeled set, $P$, that serves as augmented labeled data for training SL models alongside $L$. Instances to be placed into $L$ are selected via an AL strategy. In addition to these four sets, a set of synthetic samples, $S$, is generated for user feedback incorporation into SL model after completing each batch of explanation-feedback interactions with the user. How the steps generate/update the defined sets is defined with subscripts for interaction epochs $t \in [0, \infty]$ as follows. 

\begin{itemize}
    
    \item \textbf{Step 1: Generating a Set of Diverse Samples}
    A diverse set of items $U'$ out of $U$ is selected by UL model. 
    
    \item \textbf{Step 2: Initial Sampling to be Labeled} A diversity-based sampling strategy generates a small set $L_0$ out of $U'$. This diverse set of items is presented to the user to label them at the initial interaction step ($t=0$).

    \item \textbf{Step 3.1: Labeled Data Augmentation} Instances in $U'_t \setminus L_t$ are labeled through an SSL model based on the labels from $L_t$. Subsequently, a pseudo-labeled set $P_t$ is formed, consisting of instances with high-confidence labels assigned by the SSL model.
    
    \item \textbf{Step 3.2: Synthetic Data Generation}
    A set of synthetic samples $S^*_t$ is generated according to the user feedback (i.e., true item labels and preference explanations) in response to the system's explanations for $L'_t$. The current synthetic set is updated as $S_t = S_{t-1} \cup S^*_t$. Note that $S_0 = \emptyset$ as the user is not inquired about feedback on $L_0$ at Step 2. 
    
    \item \textbf{Step 4: Preference Model Training}
    After completing each batch of interactions (acquiring a batch of labeled samples) and applying Step 3.1 and Step 3.2, SL model is trained with $X_t = L_t \cup P_t \cup S_t$.
    
    \item \textbf{Step 5: Explanation-Feedback Interactions}
    The following explanation-feedback interaction with the user is iteratively conducted for a batch of instances $L^{*}$ selected out of $U'_t$ by an AL strategy. After completing the batch, user feedback is used to update the user preference model by applying Step 3 and 4 for $L_t = L_{t-1} \cup L^{**}$ and $U'_t = U'_{t-1} \setminus L^{**}$, where $L^{**}$ is the user-labeled instances in return to $L^{*}$.
    \begin{itemize}
        \item An instance $x$ from $L^{*}$ with its predicted label $y$ by a SL model and its explanation $E^{*}$ are presented to the user.
        \item The user responds to the system by providing a true label and true preference explanation $E^{**}$ for $x$.
    \end{itemize}
    
\end{itemize}

In the following sections, we first present how the explainable recommendation is applied in this framework. Lastly, we describe several AL strategies deployed in our study. 

\subsection{Explainable Preference Modeling}

A facet contributing to the efficiency of information exchange between the user and the recommendation system is the method of expressing preferences for items. Item rating and binary evaluation (liking or disliking) are prevalent approaches in user preference expression \cite{Ricci2015}. As detailed in Section \ref{sec-related-work}, preference expressions involving multi-point scales might be less dependable. Thus, we characterize the user preference expression as a binary evaluation of items and deploy binary classifier models for preference modeling. Furthermore, class probabilities yielded by a binary classifier can be harnessed for item ranking.

As a proof of concept for the framework, we choose to implement logistic regression (LR) as a model that can be updated quickly and is self-interpretable. Furthermore, the deployment of black-box models with an explainer method could be possible, such as a multi-layer perceptron (as the preference model) coupled with Shapley Additive Explanations (SHAP) \cite{Lundberg2017} (as the explainer method) for quantifying feature effects on general preference. However, this more complex setup necessitates sufficient time for model updates and the computation of Shapley values following each batch of user interactions.
On the other hand, LR can be easily interpreted with its coefficients which directly represent how each feature affects the estimated general preference of the user. Accordingly, we define the form of explanations derived from the deployed binary classifier as follows.

\subsection{Feature Effect-Based Preference Explanation}

Acknowledging the need for minimal cognitive exertion without compromising feedback quality in machine teaching \cite{Liao2021}, our objective in the form of preference explanation is to enhance the user's feedback experience while utilizing this feedback to personalize the recommender model. Consequently, we introduce a simple procedure for users to communicate their preferences while comprehending how the recommender model assesses items in alignment with their preferences.
To facilitate the exchange of information within the recommendation domain between the system and the user, we categorize information representing the user preference based on the effects of item features on their preferences, aligning with the explanation structure that reveals the system's recommendation rationale.  

For given coefficients of LR or feature influence scores by an explainer method, each feature is labeled based on its coefficient sign (i.e., positive or negative). If the absolute value of a coefficient is less than $\epsilon$, then that feature is labeled as `ineffective'.
Accordingly, we define the preference explanation $E^{*}_i$ for item $i$ as the categorization of its features $d \in D_i$ into positive, negative, and ineffective sets with respect to user preference: $E^+_i$, $E^-_i$, and $E^0_i$, respectively. If a feature positively influences the likability of an item in the preference model, it is included in $E^+_i$, while features with negative impact are placed in $E^-_i$. Lastly, the features that have negligible or no effect on the preference model are included in $E^0_i$. 

By presenting the categorization of features with this structure as the explanation associated with the presented or recommended item, the system asks the user to validate it by providing true categorization of features of the item. Subsequently, this user feedback is utilized by the system as follows.

\subsection{Incorporation of User Feedback into the Preference Modeling}

We propose a model-agnostic information injection technique to personalize preference models. Basically, feature labels exchanged through user interactions are added as synthetic data to the training set. In this way, ML models can learn whether each feature affects user preference positively or negatively.

The user supplies the set of correct feature labels ($E^{**}_i$) according to their general preferences for each feature in $D_i$ (the features contained in item $i$) in response to the system explanation $E^{*}_i$. 
For a given item, for each feature $ d\in D_i$, one synthetic item that has only that feature $d$ is created. For example, in a food recommendation domain, if the user gives the feedback $ \{ tomato:positive, onion:negative, pepper:ineffective \}$ for a recipe containing only tomato, onion, and pepper, then the following synthetic item vectors $s_0$ (positive instance) and $s_1$ (negative instance) with the labels $y_0 = 1$ and $y_1 = 0$, respectively, are generated:
\begin{equation*}
  s_0 =
  \begin{dcases*}
    1 ,&if d = `tomato',\\
    0, &otherwise.
  \end{dcases*}
  \qquad
  s_1 =
  \begin{dcases*}
    1 ,&if d = `onion',\\
    0, &otherwise.
  \end{dcases*},
\end{equation*} 
where $d \in D$ is the corresponding feature in the item vector.
Those synthetic items are added to $S$ with their labels. Besides, the corresponding dimension of item vectors for `pepper' is masked during the training SL model so that `pepper' has no effect on the preference model. These incrementally accumulating synthetic data throughout user interactions are used to train preference models alongside real (user-labeled) and pseudo-labeled data. To strengthen the effect of synthetic data in the training, they are weighted in the loss function of ML model. The weights of synthetic data in the loss function should be determined considering the size of real labeled data in the training set. 

\subsection{Active Learning Strategies}
\label{sec-al-strategies}
We implement a variety of AL strategies for benchmarking. In general, they are classified as diversity-based, uncertainty-based, and combined AL strategies, which take advantage of the previous two. The five implemented sampling strategies are described below, and readers can refer to \cite{Monarch2021} for further details about them. 

\begin{itemize}

    \item \textbf{Uncertainty-based Sampling} selects the most uncertain samples based on the uncertainty scores of model outputs.
    
    \item \textbf{Diversity-based Sampling} splits data into clusters and selects samples from each cluster which can be cluster centroid, cluster outliers, and randomly selected instances. The sample size from each cluster is adjusted to be proportional to the cluster size.
    
    \item \textbf{Clustered Uncertainty-based Sampling} applies diversity sampling and selects the most uncertain samples from each cluster.
    
    \item \textbf{Uncertainty-based Clustering Sampling} selects the most uncertain samples as a subset, then applies diversity sampling from the selected subset. 
    
    \item \textbf{Most Uncertain Cluster Sampling} selects all samples from the cluster with the highest average uncertainty score. If the sample size is larger than the most uncertain cluster size, it completes the informative set by selecting the next most uncertain cluster iteratively.
    
\end{itemize}

Note that Diversity-based Sampling is used for Step 1 and 2 in the proposed framework to select a diverse set of instances, and one of the other four AL strategies could be deployed in Step 5.
\section{Evaluation}
\label{Evaluation}

We implement our proposed methodology as a food recommendation system where users can evaluate recipes by specifying their preferences over recipe attributes. The proposed methodology is not restricted to food recommendation and could be implemented for several domains like e-commerce, book, movie \cite{Iovine2022}, and fashion recommendation in which recommendation items are either contain visual or textual  descriptive information that can be evaluated by users to express preferences.

We assess the effectiveness of the proposed preference elicitation framework through human experiments using a web interface and an empirical evaluation based on synthetic user profiles we generated for two food datasets \cite{Haussmann2019,CulinaryDB}.
We conducted human experiments where participants can interact with the system through a Web interface and give feedback on the displayed food recipes to evaluate the efficacy of the preference elicitation phase of the system (Section \ref{subsect-human-expriments}). 
As human experiments enable us to measure the system performance only over a limited number of interactions, we also conduct user interaction simulations with synthetic data to evaluate the system performance in the long run  (Section \ref{subsec-synthethic-experiment}).

\subsection{Human Experiments}
\label{subsect-human-expriments}

\begin{figure}[h]
\centering
\includegraphics[height=0.2\textwidth]{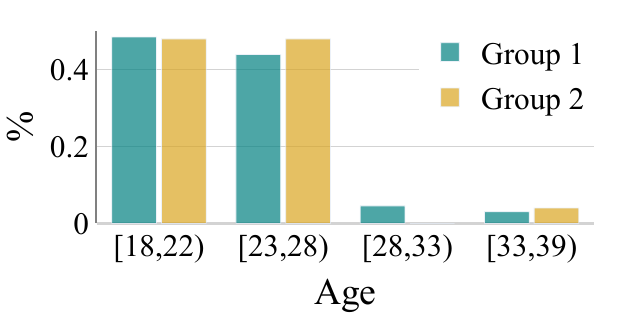}
\caption{The Age Distribution of the Participants}
\label{fig:age-distribution}
\end{figure}

By conducting online experiments with users, we aim to measure to what extent the designed preference elicitation phase is effective in terms of preference modeling accuracy at the cost of interaction burden on the users.  
98 Turkish volunteers participated in the experiment. We set two groups of participants to examine how the labeled initial sample size in Step 2 of the framework affects the trade-off between the burden of acquiring preference information from the user and the accuracy of preference modeling by the system. (i.e., the accuracy of the prediction model). 66 participants in Group 1 labeled 15 recipes in Step 2, while 25 participants in Group 2 labeled 30 recipes. More participants in Group 1 were needed to reveal an evident distribution of performance metrics as the system performance with 15 samples deviates more compared to using 30 samples in Step 2. 
The age distribution of participants in each group is depicted in Figure \ref{fig:age-distribution}. 

As food recipes from different foreign cuisines might be a barrier to explaining preferences easily and correctly for the participants, we used a dataset that includes mostly Turkish food recipes from \cite{diyetkolik} for the experiment.
The dataset includes 1854 recipes, each consisting of the following attributes (i.e., features): category, ingredients, cuisine, cooking time, calories, carbohydrates, fats, proteins, and fiber values. This dataset includes 288 different ingredients in total.
The nutritional information (carbohydrates, fat, protein, fiber, and calories) is converted to three categorical values (`Low', `Normal', and `High') based on one-third quantile values (0-0.33, 0.33-0.66, and 0.66-1.0) in the distribution of their logarithmic values. In this way, users can more easily (and consistently) specify their preferences about nutrition levels, which allows the system to learn preferences about nutritional values more accurately. In this way, recipes are represented as binary vectors.

We set up a procedure comprising the following four phases for the human experiments with a web interface:
\begin{itemize}
    \item Initial Sample Labeling (Phase 1): The user labels 15 or 30 diverse recipes as whether they like or dislike.
    \item Explanation-Feedback Interactions (Phase 2): Five recipes are presented to the user one by one. The user labels each recipe (as like/dislike) and evaluates the system's explanations associated with the recipes by classifying the recipe attributes as positive, negative, or ineffective factors.
    \item Binary Test: The user labels 25 recipes as like or dislike, where five of them are already shown in Phase 1.
    \item Rating Test: The user indicates how s/he agrees with the system's binary predictions (`Like' or `Dislike') for ten recipes as `Strongly disagree', `Disagree', `Neutral', `Agree', and `Strongly agree'. 
\end{itemize}

We used K-Medoids Clustering as the diversity sampling strategy in Phase 1 and Clustered Uncertainty-based Sampling in Phase 2. The similarities between food recipes considering the attributes are measured with Jaccard similarity \cite{Jaccard1901}.

\begin{figure}[ht]
\centering
\includegraphics[width=\textwidth]{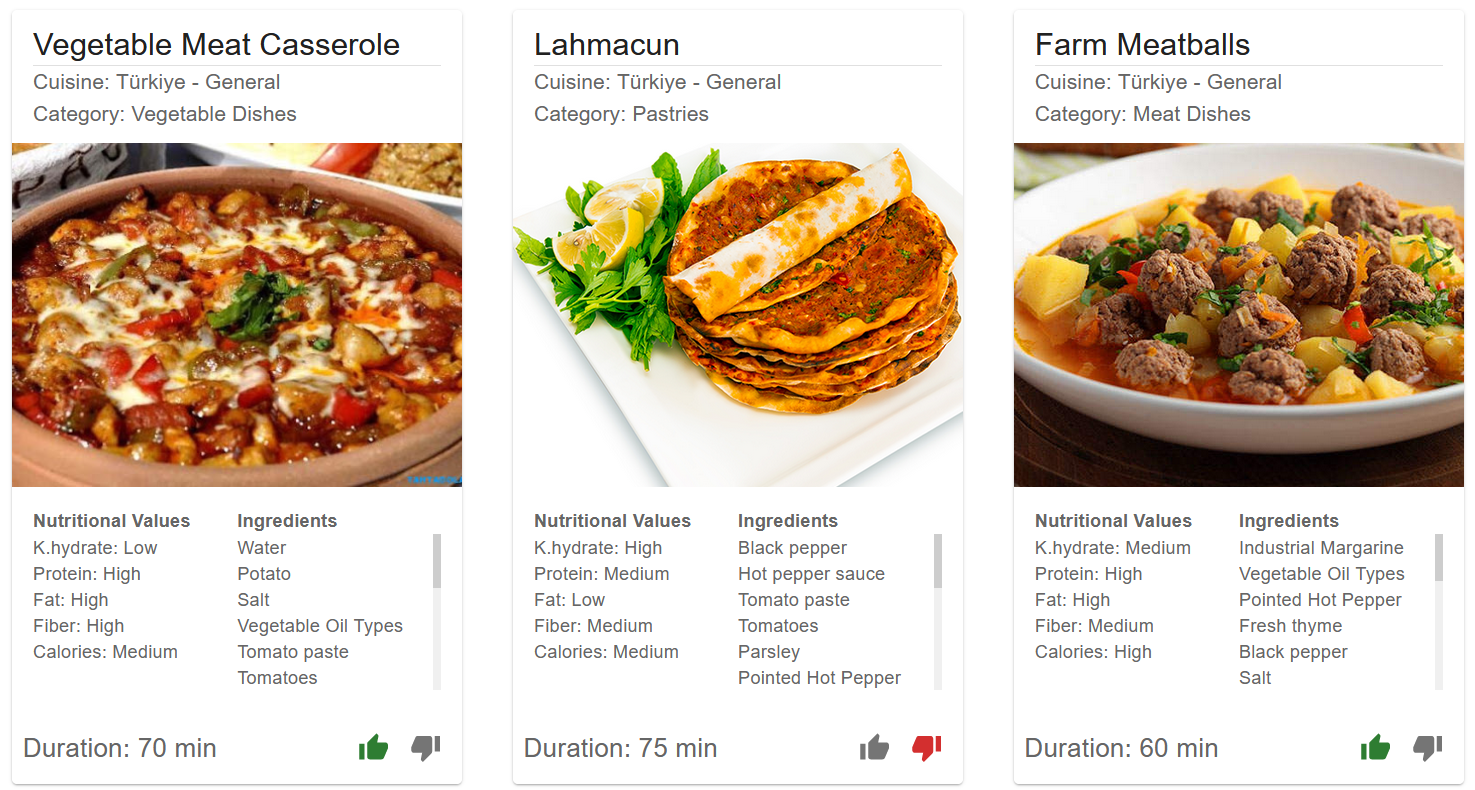}
\caption{A Part of the Preference Elicitation Interface (Phase 1)}
\label{fig:phase1_en}
\end{figure}

\begin{figure}
\centering
\includegraphics[width=\textwidth]{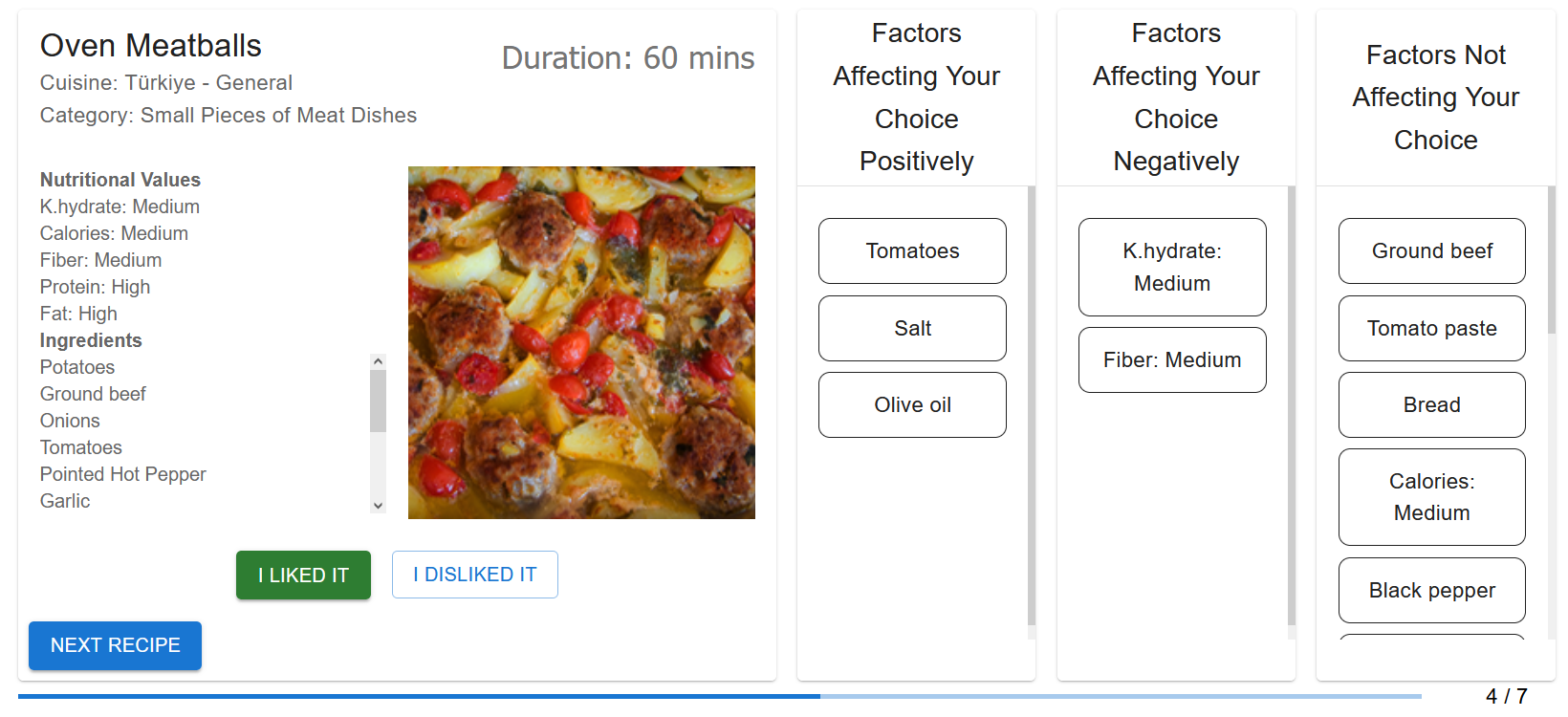}
\caption{The Fine-Grained Preference Elicitation Interface (Phase 2)}
\label{fig:phase2_en}
\end{figure}

As shown in Figure \ref{fig:phase1_en}, users are asked to indicate whether they like or dislike the displayed food recipes in Phase 1.
In the next phase, users are asked to specify structured explanations for their preferences on a chosen recipe by indicating the factors (i.e., ingredients, cooking time, cuisine, and nutrition levels) affecting their choices positively and negatively, as well as the factors not having any influence on their choice as seen in Figure \ref{fig:phase2_en}. Since this process may increase the participants' cognitive load significantly, they are inquired to evaluate only five recipes in total. The preference model is updated after receiving user input for each recipe in Phase 2. We provide the implementation details of the Web interface in Appendix \ref{sec:web-interface}.

The third and fourth phases evaluate the preference model learned after Phases 1 and 2. 
Firstly, 30 recipes are selected using a testing sampling approach, which is detailed in Appendix \ref{sec:test-sampling}. It selects the instances where the model predictions are the most certain and then discards similar instances.
In Binary Test, the system asks users to indicate whether they like or dislike 20 testing recipes through the same interface as in Phase 1. Additionally, five recipes from Phase 1 are presented in Binary Test to check the users' consistency. 
If the user labels at least two recipes differently than those in Phase 1, we do not consider those users' results in the evaluation. 
In Rating Test, the system presents ten food recipes to the user with the predicted like/dislike labels.
The users specify to what extent they agree with the system decision on a 5-point likert scale (i.e., strongly/moderate disagree, neutral, moderate/strongly agree).

The main difference between Binary Test and Rating Test, other than labeling types, is that the participants are unaware that they score the system performance in Binary Test step because any instruction specifying that the labels given by the users to be used to evaluate the performance of the system was not provided to them in the experiment. On the other hand, the participants are explicitly inquired to evaluate the system's predictions in Rating Test step, and they express how they agree with the system's like/dislike prediction for each recipe with one of the responses. Resorting to both types of testing, we can evaluate the system performance more reliably against two possibilities of misinformation reveal. Firstly, label noise might yield from Binary Test because people may provide inattentive or inconsistent labels for some recipes out of 25 due to feeling exhausted from labeling a high number of items. Secondly, in Rating Test people might evaluate the experimented system with over-appreciation that could yield more positive bias towards the system's predictions, conversely, or more negative bias due to some opposition towards the experiment. We also included five random samples from Phase 1 in Binary Test to check whether participants labeled the recipes carefully. We excluded the results of 7 participants who labeled at least two recipes in Binary Test differently than in Phase 1.  

Before starting the experiments, the experiment protocol adopted in this study was approved by the Ethics Committee of our university, and informed consent was obtained from all participants. In addition, they watched a short video demonstrating how they should interact with the system before each phase in the experiment. 

\subsubsection{Experimental Results}
\label{sec-human-results}

\begin{figure}[h]
    \begin{subfigure}[b]{0.49\textwidth}
    \centering
    \includegraphics[width=\textwidth]{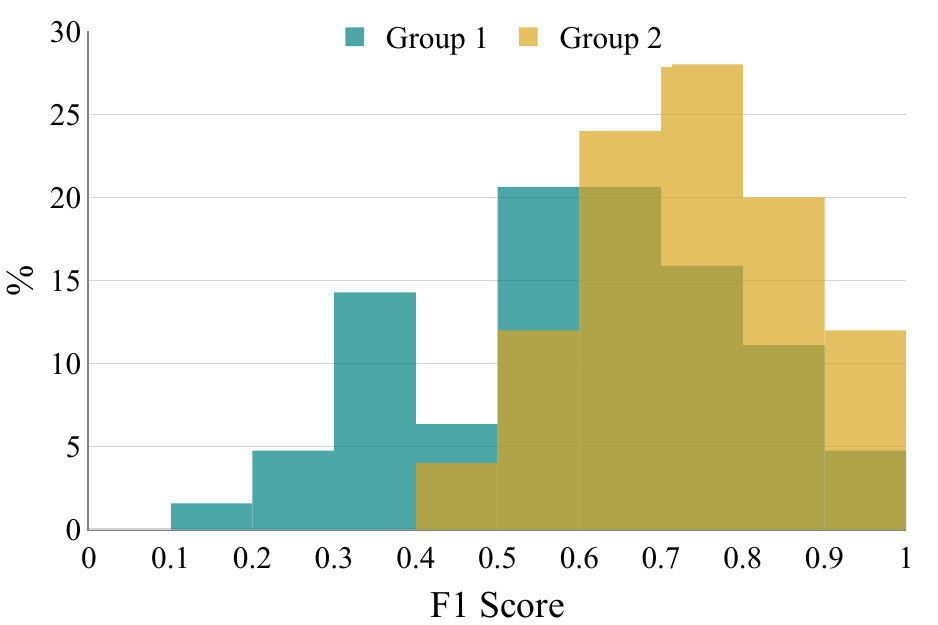}
    \caption{F1 Scores on Binary Test Samples by the Experiment Groups}
    \label{fig:f1_distribution}
    \end{subfigure}
    \hfill
    \begin{subfigure}[b]{0.49\textwidth}
    \centering
    \includegraphics[width=\textwidth]{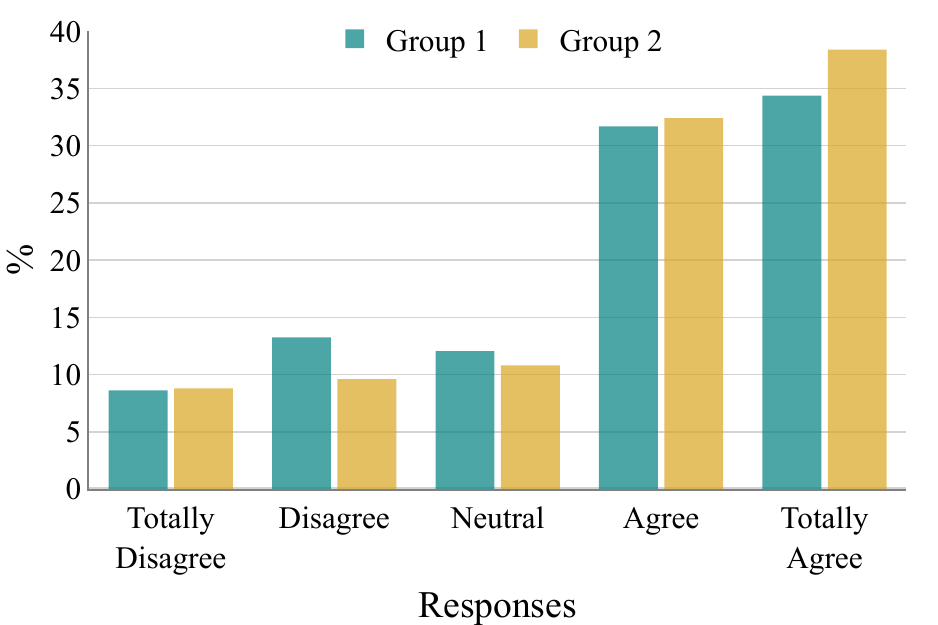}
    \caption{Responses for Rating Test Samples by the Experiment Groups}
    \label{fig:rating_distribution}
    \end{subfigure}
\caption{The Testing Results of Human Experiments}
\end{figure}

Figure \ref{fig:f1_distribution} presents the distribution of the average F1 scores of the recommender models in Binary Test. We applied the two-tailed Mann-Whitney U test \cite{Mann1947} for two groups' scores, which do not have a homogeneous variance, that validates the significant difference ($p <0.015$) between the groups. The score distribution of Group 1 lies between 0.143 and 1.0 F1 with an average of 0.612, while Group 2's results lie between 0.43 and 0.974 F1 with an average of 0.725. It is revealed that labeling more recipes in the preference elicitation phase (30 versus 15) improves the capability of the system's preference modeling by 18.5\% more F1-score on average.  
Although scoring the system with the participants' labels in Binary Test implicitly prevents any bias toward the system performance, users might not be careful when labeling 25 samples as a batch, which may produce label noise. Therefore, we also evaluate the system performance differently with Rating Test, in which users explicitly score the system.
Figure \ref{fig:rating_distribution} presents percentages of responses within each group in Rating Test. The significant difference between the groups is seen in the percentages of ``Disagree'' and ``Totally Agree'' responses. Group 1 disagrees with the system prediction at 13.24\% while this rate decreases to 9.6\% in Group 2. Another evidence of refinement of the preference modeling with 15 more labeled recipes is that 38.34\% of Group 2 responses are ``Totally Agree'' while it is 34.37\% for Group 1.

%%%%%%%%%%%%%%%%%%%%%%%%%%%%%%%%%%%%%%%%%%%%%%%%%%%%%%%%%%%%%%%%%%%%%
%%%%%%%%%%%%%%%%%%%%%%%%%%%%%%%%%%%%%%%%%%%%%%%%%%%%%%%%%%%%%%%%%%%%%

The participants answered the following 5-point likert scale survey questions at the end of the experiment. These questions measure the system in terms of \textit{labeling burden}, \textit{feedback burden}, \textit{feedback complexity}, and \textit{explainability}, respectively. Here, we use the terms  ‘burden' and  ‘complexity' to distinguish between the quantitative and qualitative aspects of user interaction load, respectively.

\begin{enumerate}
\item How exhausting was it to label 15/30 recipes in Phase 1? (1: Slightly, 5: Too much)
\item How exhausting was it to repeat Phase 2 for five recipes? (1: Slightly, 5: Too much)
\item How exhausting was it to classify food attributes as positive/negative/ineffective based on your food preferences in Phase 2? (1: Slightly, 5: Too much)
\item How clear the classification of food attributes as positive/negative/ineffective factors to explain your food preference is? (1: None, 5: Totally)
\end{enumerate}

Table \ref{tab:survey-results} provides the scaled average values of the scores (between 0 and 1) given by the participants in this survey. The first column of the table is the performance criteria corresponding to the questions. The next two columns are average scaled scores within each group, while the fourth column includes the scaled scores averaged over all participants. Although it is not necessary for the criteria other than \textit{labeling cost}, we applied the Mann-Whitney U-Test to certify no significant divergence between the mean scores given by the two groups. The last column of Table \ref{tab:survey-results} includes $p$ values for the one-tailed test for Labeling Burden and the two-tailed test for others. We see that the participants in Group 2 found Phase 1 more exhausting than Group 2 ($p \leq 0.032$). Although participants in Group 2 label the doubled number of recipes, they find Phase 1 34\% more exhausting on average compared to Group 1.  While expressing food preferences with binary labels for a list of recipes at once is not much frustrating for the participants, they feel slightly more burdened on themselves when providing feedback for five recipes one by one. Nevertheless, limiting the number of recipes to get user feedback to 5 yields 0.35 on average. Another interaction cost comes from the complexity of explaining food preferences. The survey result shows that participants find categorizing item attributes (as a preference explanation form) exhausting, with a score of 0.4374 on average. In return for this, the system provides a much higher score in terms of explainability, which is 0.725 on average.

%%%%%%%%%%%%%%%%%%%%%%%%%%%%%%%%%%%%%%%%%%%%%%%%%%%%%%%%%%%%%%%%%%%%%
\begin{table}[h]
\caption{The Normalized Average Scores by Two Experiment Groups in the Survey}
\label{tab:survey-results}
\centering
\begin{tabular}{lcccc}
\toprule
                               & \textbf{Group 1} & \textbf{Group 2} & \textbf{All} & \textbf{$p$-value} \\ \hline
\textbf{Labeling Burden}     & 0.238            & 0.340            & 0.268        & 0.032            \\ 
\textbf{Feedback Burden}     & 0.323            & 0.350            & 0.330        & 0.482            \\ 
\textbf{Feedback Complexity} & 0.435            & 0.390            & 0.423        & 0.442            \\ 
\textbf{Explainability}        & 0.728            & 0.710            & 0.723        & 0.893            \\ 
\bottomrule
\end{tabular}
\end{table}
%%%%%%%%%%%%%%%%%%%%%%%%%%%%%%%%%%%%%%%%%%%%%%%%%%%%%%%%%%%%%%%%%%%%%

\subsection{Synthetic User Experiments}
\label{subsec-synthethic-experiment}

For empirical evaluation of our XAL approach for user preference elicitation in the food recommendation domain, we constructed a procedure to establish synthetic user profiles consisting of scores for recipe ingredients and recipe labels. Nutrition values, cuisine, cooking time, recipe ingredients, and food category are some attributes of food datasets, but we only use ingredients as recipe attributes to construct synthetic user profiles, considering them as the most common factor in food ratings according to the survey in \cite{Harvey2012}. Also, by using only one feature set, we can prevent causing inconsistencies in the preference of the synthetic profiles.

In this section, we present the results for FoodKG dataset \cite{Haussmann2019}, which includes 69265 recipes and 343 ingredients after data processing. Besides, the results for CulinaryDB dataset \cite{CulinaryDB} are provided in Appendix \ref{sec:culinarydb_results}, drawing similar conclusions.
To generate synthetic profiles, we manually categorized the 200 most frequent ingredients, with the lowest having a frequency of 0.8\% in the dataset, into 22 groups (e.g., meaty, seafood, vegetable, carby, eggy, milky, spicy, sweet, etc.)  considering taste, nutrition, main substance, food type, healthiness, and meal context. The remaining 143 ingredients were left ungrouped. By assigning scores to relevant ingredient groups, we created four eating style templates (vegan, sportive, elder, and unhealthy) and three user templates, each based on real personal preferences. The scores were semantically varied for each type of template. 
For example, a vegan profile template was generated by firstly scoring ingredients in the following groups with the corresponding score ranges: $ \{ \text{vegetable}:[8,10], \text{fruit}:[8,10], \text{green}:8, \text{meaty}:[1,1], \text{eggy}:[1,1], \text{milky}:[1,1] \} $. Then, the rest ingredient groups were scored randomly in $[1, 10]$, where the lowest preference is one and the highest is ten.  
Furthermore, for each profile, randomly selected 20-30\% of the ingredients were retained as ineffective recipe attributes in food preference. 
Accordingly, we set nine different template profiles by combining the eating style preferences and the real preferences, along with a random profile in which all ingredient scores are randomly assigned. Eventually, using these ten template profiles, we created 50 distinct synthetic profiles by altering random seeds five times to assign ingredient scores. Then, we define the recipe score function as the mean of the scores of the effective attributes while excluding the ineffective ones from the calculation.

We establish a rule set for providing feedback through synthetic user profiles in response to the system's explanations for recipe preferences. Explanation-feedback interactions are simulated according to the following synthetic user rationale.

\begin{itemize}
    \item \textbf{Recipe Labeling}: A recipe is liked if its score exceeds a threshold of 5.5 (out of 10); otherwise, it is disliked. 
    \item \textbf{Feature Labeling}: If an ingredient belongs to the set of ineffective features, it is labeled as an ineffective factor in recipe preference. Otherwise, it is classified as a positive (negative) factor influencing the likability of the recipe if the ingredient score is higher than (lower than or equal to) the average of recipe scores.
\end{itemize}

For the preference elicitation phase, we set the initial sample size $|L_0|$ to 30, the batch size (i.e., the number of recipes selected by AL at a time) to one, and the number of user interactions to five. After getting user feedback for each recipe, the preference model is updated by retraining, and a new instance is selected with a given AL strategy. 
After this preference phase of five interactions, we also simulate 45 interactions to analyze how effective the proposed solution is in the refinement of preference models if it is deployed throughout the recommendation phase of the system. This envisaged recommendation phase includes asking the user to provide recipe labels (like or dislike) and feedback about the system's explanation for a batch of recipes selected with AL. After each batch, the preference model is updated, and a new batch to be labeled is supplied to the system for the next interaction. For this long-run analysis covering 45 interactions, we set the batch size to 5. Therefore, 5+45 interactions are conducted with the synthetic user profiles for 5+225 recipes in total. 

\subsubsection{Empirical Results}

\begin{figure} [hb]
\centering
\includegraphics[height=0.4\textwidth]{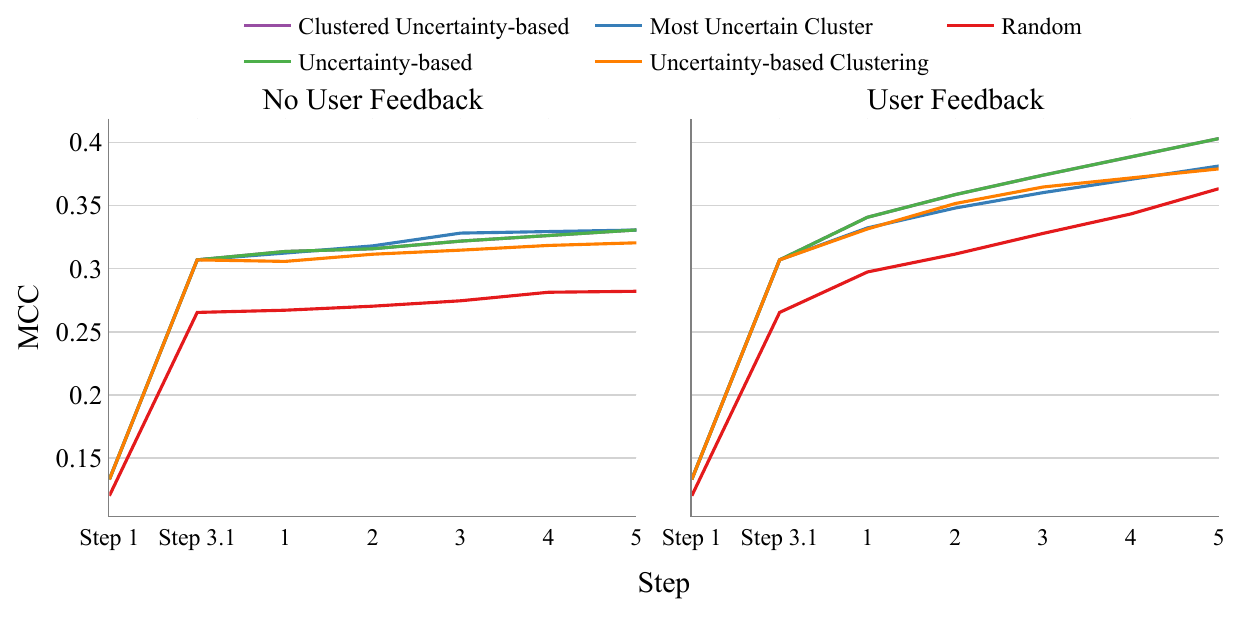}
\caption{The Average MCC without (left) and with (right) User Feedback in the Preference Elicitation Phase for FoodKG Dataset}
\label{fig:foodkg-preference-elicitation}
\end{figure}

To assess performance gain by AL strategies for eliciting user preferences accurately, they are scored over unlabeled data left ($U_t$) after completing each batch of interactions with the user. 
We also provide results by random sampling to reveal how the proposed framework benefits from AL. 
As we aim to examine the overall success of preference modeling with binary classifiers (regarding both `like' and `dislike' classes) in the long run analysis, we report Matthews correlation coefficient (MCC $\in [-1,1]$) \cite{Matthews1975} 
We consider the accuracy of the preference explanation by the trained model after each interaction step, which is calculated as the ratio of true feature labels. In each result analysis, scores are averaged over unspecified dimensions in that analysis. Table \ref{tab:0to50-performance} includes the average MCC by AL strategies for the first five interactions, which are visualized in Figure \ref{fig:foodkg-preference-elicitation}, and the last interaction step (t=50) with the breakdowns for whether user feedback is incorporated into the models. The improvement by AL strategies over Random Sampling is notable in the first five steps. Moreover, the value of AL becomes more prominent in the long run, and they can acquire 6\% more MCC at t=50. When recipes for the diverse set ($U'$) and $L_0$ are selected randomly, the preference models achieve 0.124 and 0.218 MCC on average after training them with the initial labeled sample ($L_0$) in Step 2 and the augmented labeled data ($L_0 \cup P_0$) in Step 3.1, respectively. On the other hand, the system acquires preference models providing 0.145 and 0.272 average MCC when Diversity Sampling is used in Step 1 and Step 2. 87.5\% MCC improvement through Step 2 shows the solid contribution of SSL for preference elicitation with limited labeled data in the early phase.

\begin{table}[h]
\centering
\caption{The Average MCC by LR at  $t \in \{1, \ldots, 5 \}$ and $t=50$ for FoodKG Dataset}
\label{tab:0to50-performance}
%\resizebox{\columnwidth}{!}{%
\begin{tabular}{cccccccc}
\toprule
\textbf{\begin{tabular}[c]{@{}c@{}}Sampling\\ Strategy\end{tabular}} &
  \textbf{\begin{tabular}[c]{@{}c@{}}User \\ Feedback\end{tabular}} &
  \textbf{t=1} &
  \textbf{t=2} &
  \textbf{t=3} &
  \textbf{t=4} &
  \textbf{t=5} &
  \textbf{t=50} \\ \hline

\multirow{2}{*}{\begin{tabular}[c]{@{}c@{}}Clustered\\ Uncertainty-based (\texttt{CU})\end{tabular}}  & Not Used & 0.314 & 0.316 & 0.322 & 0.326 & 0.331 & 0.593 \\ 
                                                                                        & Used     & 0.341 & 0.359 & 0.374 & 0.388 & 0.403 & 0.709 \\ \hline
\multirow{2}{*}{Uncertainty-based (\texttt{Ub})}                                                      & Not Used & 0.314 & 0.316 & 0.322 & 0.326 & 0.331 & 0.587 \\
                                                                                        & Used     & 0.341 & 0.359 & 0.374 & 0.388 & 0.403 & 0.710 \\ \hline
\multirow{2}{*}{\begin{tabular}[c]{@{}c@{}}Uncertainty-based\\ Clustering (\texttt{UC}\end{tabular}} & Not Used & 0.306 & 0.311 & 0.315 & 0.318 & 0.321 & 0.526 \\  
                                                                                        & Used     & 0.332 & 0.352 & 0.365 & 0.372 & 0.379 & 0.656 \\ \hline
\multirow{2}{*}{\begin{tabular}[c]{@{}c@{}}Most Uncertain \\ Cluster (\texttt{MUC})\end{tabular}}      & Not Used & 0.312 & 0.318 & 0.328 & 0.329 & 0.331 & 0.538 \\  
                                                                                        & Used     & 0.332 & 0.348 & 0.360 & 0.370 & 0.381 & 0.699 \\ \hline
\multirow{2}{*}{Random}                                                                 & Not Used & 0.267 & 0.270 & 0.275 & 0.281 & 0.282 & 0.466 \\  
                                                                                        & Used     & 0.297 & 0.312 & 0.328 & 0.343 & 0.363 & 0.643 \\                                                                                 
\bottomrule
\end{tabular}%
%}
\end{table}

%%%%%%%%%%%%%%%%%%%%%%%%%%%%%%%%%%%%%%%%%%%%%%%%%%%%%%%%%%%%%%%%%%%%%
%%%%%%%%%%%%%%%%%%%%%%%%%%%%%%%%%%%%%%%%%%%%%%%%%%%%%%%%%%%%%%%%%%%%%

Note that each AL strategy starts with the same preference model, which was obtained with the training after Diversity Sampling in Step 1 and 2, for the first explanation-feedback interaction; therefore, their performance is equal at the end of Step 1 and Step 3.1. On the other hand, Random Sampling uses the model obtained with Random Sampling in Step 1 and 2, resulting in lower MCC values.
Monotonic increases in the performance of AL strategies are seen in the first five explanation-feedback interactions with the synthetic users. Moreover, incorporating user feedback into the preference models provides 21\% more MCC on average at the end of the first five interactions. The best-performing AL strategies, Clustered Uncertainty-based Sampling (\texttt{CU}) and Uncertainy-based Sampling (\texttt{Ub}), provide 10.1\% (17.2\%) more MCC on average compared to random sampling when user feedback is used (not used). 
Note that \texttt{CU} and \texttt{Ub} strategies perform the same in the first five steps because AL strategies select only one instance (the batch size is one) at each of these steps. Therefore, both \texttt{CU} and \texttt{Ub} select the most uncertain instance.

\begin{figure}[h]
\centering
\includegraphics[height=0.4\textwidth]{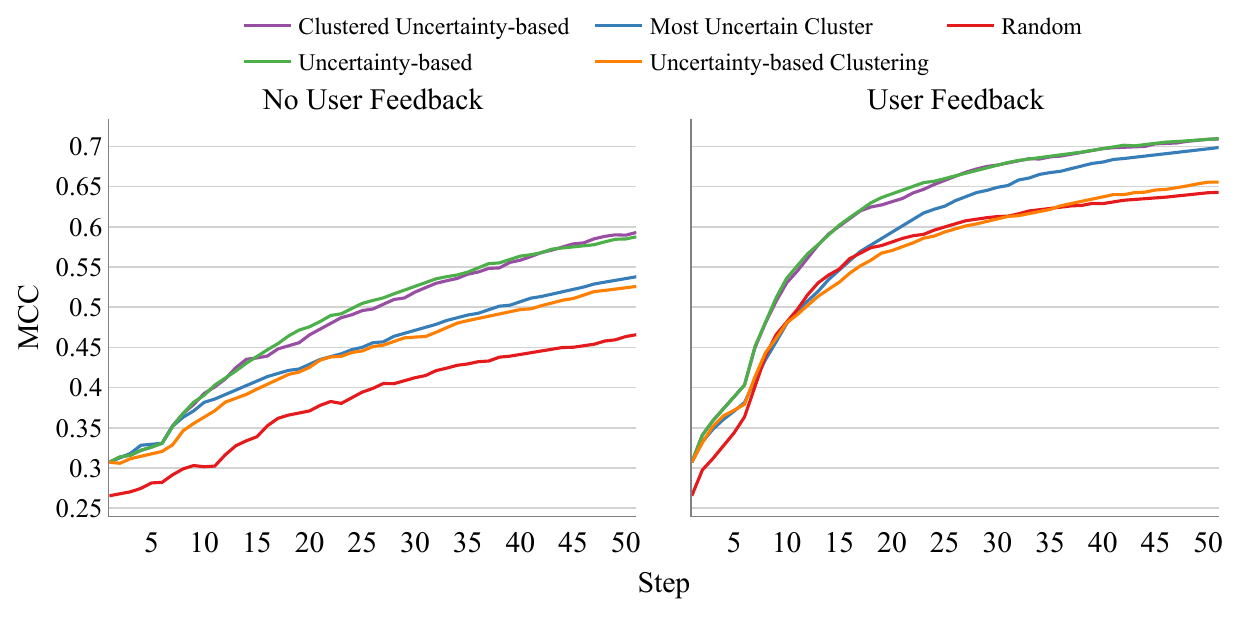}
\caption{The Average MCC by LR without (left) and with (right) User Feedback in 50 Interaction Steps for FoodKG Dataset}
\label{fig:foodkg-50-steps}

\end{figure}

Although performance gained through Step 3.1 and Step 3.2 in the first five user interactions are remarkable, the advantage of AL strategy over Random Sampling for preference elicitation is proven in the long run with Figure \ref{fig:foodkg-50-steps}. At the end of 50 steps, \texttt{CU} and \texttt{Ub}, as the best-performing AL strategies, provides 10.4\% (27.3\%) more MCC compared to Random Sampling with (without) using user feedback. To interpret the performance gain by AL strategies over Random Sampling in terms of user interaction cost, Table \ref{tab:step_difference} presents how many steps the preference model with Random Sampling can reach the average MCC level of each AL strategy. For example, if the user feedback is not used to update the preference model, \texttt{CU} achieves the average MCC value of Random Sampling at t=50 $\lfloor 30.99 \rfloor$ interaction steps earlier, which corresponds to$ 30 \times 5 = 150$ fewer recipes labeled by the user. Consequently, the results presented in this table summarize how AL benefits to realize the objective of this study, maximizing the information acquisition from the user with minimal user interaction.

It is worth noting that Jaccard similarity is about 0.1 on average in the food datasets; namely, the similarity of the recipes in the food recommendation domain is well low. Hence, it is observed through Figure \ref{fig:foodkg-50-steps} that \texttt{Ub} and \texttt{CU} perform similarly. If many similar instances existed in the dataset \texttt{Ub} would underperform. In that case, the uncertainty ranking includes consecutive similar instances, so selecting the top n instances with \texttt{Ub} would provide less information. However, recipes with hundred ingredients represent a well-sparse dataset. Therefore, two consecutive recipes in the ranking are unlikely to be similar.

%%%%%%%%%%%%%%%%%%%%%%%%%%%%%%%%%%%%%%%%%%%%%%%%%%%%%%%%%%%%%%%%%%%%%
%%%%%%%%%%%%%%%%%%%%%%%%%%%%%%%%%%%%%%%%%%%%%%%%%%%%%%%%%%%%%%%%%%%%%

\begin{figure}[h]
\centering
\includegraphics[height=0.4\textwidth]{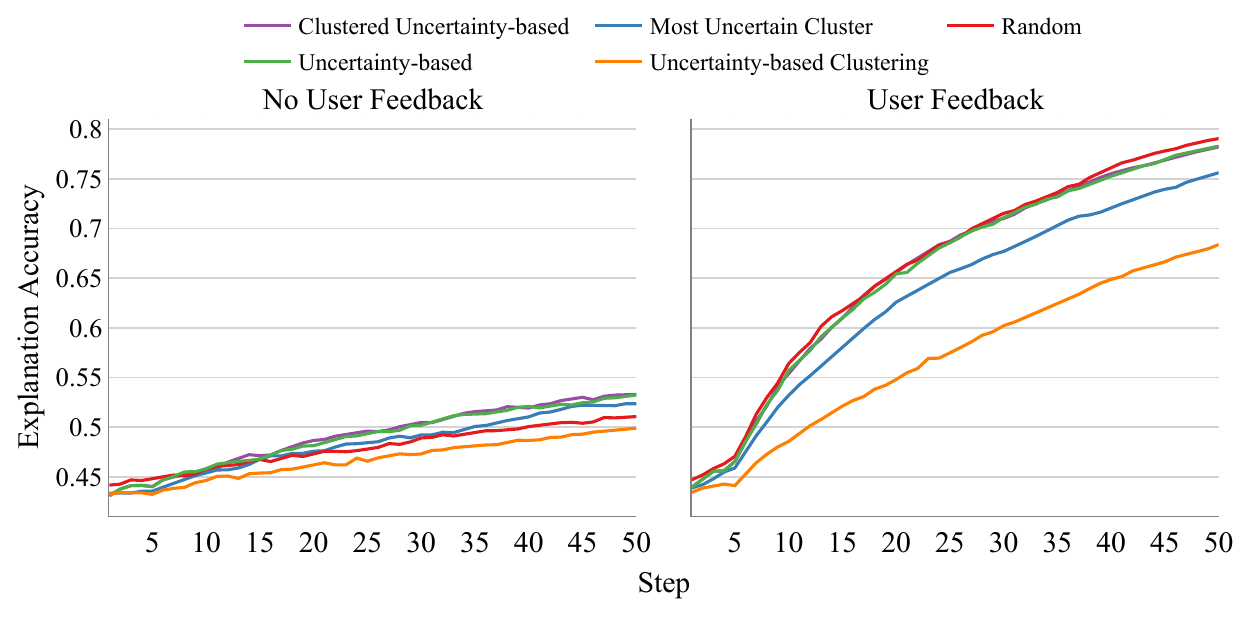}
\caption{The Average Explanation Accuracy without (left) and with (right) User Feedback in 50 Interaction Steps for FoodKG Dataset}
\label{fig:foodkg-explanation-acc}
\end{figure}

Lastly, Figure \ref{fig:foodkg-explanation-acc} illustrates the average evolution of preference explanation accuracy over the steps and the advantage of the proposed approach to incorporating user feedback into ML. Utilizing user feedback to update preference models improves the explanation accuracy by a 50-79\% increase at t=50. Interestingly, preference models obtained through random sampling exhibit a greater improvement in explanation accuracy through user feedback compared to AL strategies for FoodKG dataset. This discrepancy arises from the fact that AL methods aim to gather information pertinent to instance labels, whereas the proposed user feedback scheme also supplies labels for features of items. On the other hand, the explanation accuracy achieved by the models of \texttt{CU} and \texttt{Ub} is generally on par with that of Random Sampling for CulinaryDB dataset (refer to Figure \ref{fig:culinarydb-explanation-acc}). 
These results offer a complementary perspective on how the proposed XAL framework functions for preference elicitation. The explainability of the system facilitates the enhancement of the trained model's reasoning regarding the impact of recipe features on the user's preference. This is achieved through the acquisition of feature labels from the user, prompted by the system's explanations. Additionally, the explainability aspect contributes to the overall success of preference elicitation.

\begin{table}[h]
\centering
\caption{The Horizontal Distance Between the Average (Interpolated) MCC Curve of Random Sampling and That of Each AL Strategy in Figure \ref{fig:culinarydb-50-steps}}
\label{tab:step_difference}
\begin{tabular}{cclllll}
\toprule
\textbf{\begin{tabular}[c]{@{}c@{}}Sampling\\ Strategy\end{tabular}} &
  \textbf{\begin{tabular}[c]{@{}c@{}}User\\ Feedback\end{tabular}} &
  \multicolumn{1}{c}{\textbf{t=10}} &
  \multicolumn{1}{c}{\textbf{t=20}} &
  \multicolumn{1}{c}{\textbf{t=30}} &
  \multicolumn{1}{c}{\textbf{t=40}} &
  \multicolumn{1}{c}{\textbf{t=50}} \\ \hline
\multirow{2}{*}{\begin{tabular}[c]{@{}c@{}}Clustered\\ Uncertainty-based (\texttt{CU})\end{tabular}}  & Not used & 10.03 & 12.1  & 18.7  & 24.56 & 30.99 \\ 
                                                                                        & Used     & 2.39  & 7.41  & 14.73 & 20.99 & 28.86 \\ \hline
\multirow{2}{*}{\begin{tabular}[c]{@{}c@{}}Most\\ Uncertain Cluster (\texttt{MUC})\end{tabular}}       & Not used & 10.03 & 11.36 & 14.67 & 17.77 & 22.42 \\ 
                                                                                        & Used     & -0.12 & 1.89  & 8.5   & 15.17 & 22.91 \\ \hline
\multirow{2}{*}{Uncertainty-based (\texttt{Ub})}                                                      & Not used & 10.03 & 12.28 & 18.62 & 25.49 & 32.79 \\  
                                                                                        & Used     & 2.49  & 7.37  & 14.8  & 22.77 & 30.64 \\ \hline
\multirow{2}{*}{\begin{tabular}[c]{@{}c@{}}Uncertainty-based\\ Clustering (\texttt{UC} \end{tabular}} & Not used & 10.03 & 9.37  & 13.24 & 17.1  & 19.58 \\  
                                                                                        & Used     & -0.52 & -1.96 & -0.38 & 3.09  & 7.48  \\ 
                                                                                        
\bottomrule                                                                                               
\end{tabular}
\end{table}

%%%%%%%%%%%%%%%%%%%%%%%%%%%%%%%%%%%%%%%%%%%%%%%%%%%%%%%%%%%%%%%%%%%%%
%%%%%%%%%%%%%%%%%%%%%%%%%%%%%%%%%%%%%%%%%%%%%%%%%%%%%%%%%%%%%%%%%%%%%

\section{Conclusion}
\label{sec-conclusion}

Preference elicitation without exploiting a base of domain data like collaborative information and historical implicit/explicit user feedback requires an intelligent system interacting with users to elicit their preferences. Accordingly, this study proposes a preference elicitation framework maximizing acquired information from the user via operating AL within user interactions of presenting preference-based explanations in return of getting user feedback to personalize the preference model. The human experiments that we conducted and an empirical evaluation of our methodology with two food datasets demonstrate the efficiency of XAL for preference elicitation in the food recommendation domain. 

A possible extension of the proposed framework could be refining the local reasoning of the trained ML model by inquiring users to express their item-level preference rather than general preference across all domain features. For example, one could dislike pizza with pineapple, although s/he likes pizza and pineapple independently. After obtaining a "local explanation" for specific attributes by the user, it becomes necessary to update the estimated preference model in a way that refines its decision boundary for a specific region of the domain, without impairing its overall global reasoning.
Furthermore, one salient challenge of adopting AL in RSs is the prerequisite of tailored AL strategy for each domain and even for each user. As a future work, AL strategies adopting meta-learning could be designed to learn which strategy should be used for which user prototypes defined in a given recommendation domain. Additionally, the proposed methodology could be upgraded to deploy dynamic preference modeling approaches for changing user preferences in time by introducing user behavior-aware learning mechanisms.

\begin{acks}
This work has been supported by the \chistera grant, CHIST-ERA-19-XAI-005 and by the Scientific and Research Council of T\"urkiye (T\"UB\.{I}TAK, G.A. 120N680). We would like to thank Berk Buzcu for his support in the human experiments of this study.
\end{acks}

\bibliographystyle{ACM-Reference-Format}
\bibliography{references}

\appendix
\begin{appendices}

\section{Implementation Details}
This section provides the implementation details of the study.

\subsection{The Test Sampling Approach Used for the Human Experiments}
\label{sec:test-sampling}

For Binary and Rating tests, we need a recipe list that can validate how the learned preference models are accurate based on the acquired information in Phase 1 and 2. Therefore, we take the instances in which the model predictions are the least uncertain. On the other hand, test samples are required to be diverse since those confidently predicted instances can be very similar to each other, which leads to a very biased evaluation. To avoid such a biased evaluation, we iteratively discard similar instances until we get the total test sample size. Firstly, instances are sorted in the ascending order of uncertainty scores, and instances are selected in order to fill up the test sample at size 20 by iteratively discarding instances that have larger than 0.25 (which is above the average ~0.2 of within-cluster similarities) a Jaccard similarity with the instances already in the test sample. In this way, we can ensure the diversity of test samples which provides a more general scoring over the food dataset than only scoring top predictions, which can include very similar recipes. 

\subsection{Web Interface for the Human Experiment}
\label{sec:web-interface}

There are two types of pages in the web interface used for the human experiment. Figure \ref{fig:phase1_en} includes a portion of the page in which a list of recipes is presented to the user to label as `Like' or `Dislike'. Each recipe card on this page comprises the recipe attributes in Turkish, which are cuisine, category, nutritional values, ingredients, and total time (preparation and cooking). Users can label recipes with buttons on the bottom left of the cards. The second page of the web interface is presented in \ref{fig:phase2_en}, which presents only one recipe card to the user at a time. This page includes three boxes to classify recipe attributes as positive, negative, or ineffective factors on the user's food preference. In the initial state of the page, boxes for recipe attributes are placed according to the system's classification; then, users can relocate them according to their preferences. The colors of the relocated recipe attributes boxes are changed to green, red, and gray, respectively.

While interacting with users, the system displays the nutritional information (carbohydrates, fat, protein, fiber, and calories) as categorical values (`Low', `Normal', and `High'), instead of their real values, based on one-third quantile values (0-0.33, 0.33-0.66, and 0.66-1.0) in the distribution of their logarithmic values. In this way, users can more easily (and consistently) specify their preferences about nutrition levels, which allows the system to learn preferences about nutritional values more accurately.

\subsection{Synthetic User Experiments}
\label{sec:synthetic_exp_details}

There are several parameters of the framework to be set based on the data size of the application domain. In this section, we detail how we determined these parameters. 

For Steps 1 and 2, a variety of well-known clustering algorithms (Affinity Propagation, Density-based Clustering, Agglomerative Clustering, and K-Means Clustering) are benchmarked, but each approach yielded highly imbalanced cluster sizes due to the sparsity of food data that causes AL strategies perform poorly. Instead, we used K-Medoids Clustering \cite{Park2009} with Jaccard distance which yielded less imbalanced cluster sizes. For step 1, all recipes were split into 100 clusters. The size of the diverse set $U'$ sampled from those 100 clusters is set to 5000. For Step 1, those 5000 recipes were split into 30 clusters. We use the clustering fitted in Step 2 for the combined AL strategies that are deployed in Step 5 as well because it is observed that re-fitting the clustering algorithm for each strategy in each epoch might yield unstable preference models. Using the initially fitted clustering algorithm can benefit constantly improving preference models over steps. 

For Step 3.1, Label Spreading (LS) \cite{Zhou2003}, a classical SSL algorithm, is deployed using the Jaccard distance for the affinity matrix representing pairwise item similarities. Samples that are labeled by LS with high confidence are added to the supervised training set $X_t$. We set the confidence threshold as the normalized entropy value of class probabilities. 

For Step 3.2, we set the coefficient ($k$) for the loss term of synthetic samples to 20. As the implementation of Logistic Regression in Scikit-learn library does not allow the modification of the loss function, we multiplied the number of synthetic samples by 20 in the training dataset. We also tried to set $k$ dynamically for each labeled feature based on the number of training samples that contain the labeled feature.
However, this approach resulted in a lower performance gain.

We used Least Confidence as the uncertainty metric. However, ranking predictions respective to different uncertainty metrics results in the same order for binary classification; therefore, it does not matter which uncertainty metric is used for the implemented AL strategies in this study. 

\section{Results for CulinaryDB Dataset}
\label{sec:culinarydb_results}
The following results are for the synthetic user experiments with CulinaryDB dataset \cite{CulinaryDB}, which includes 38661 recipes and 291 ingredients obtained after the data processing.

%%%%%%%%%%%%%%%%%%%%%%%%%%%%%%%%%%%%%%%%%%%%%%%%%%%%%%%%%%%%%%%%%%%%%
\begin{figure}[h]
\centering
\includegraphics[height=0.4\textwidth]{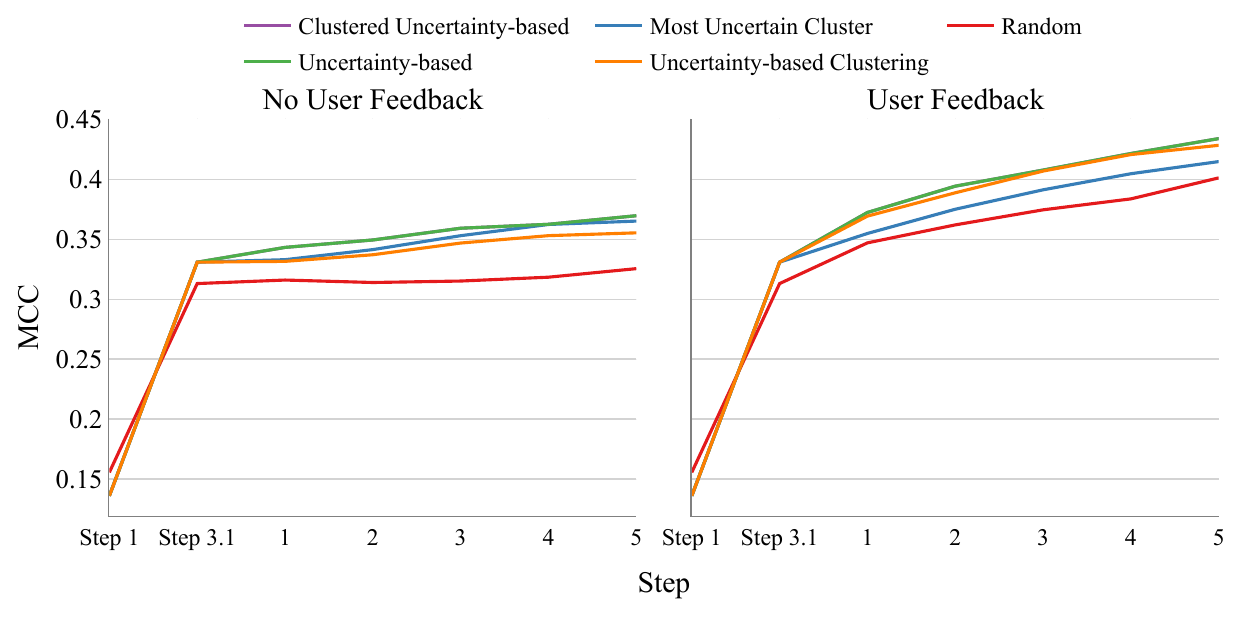}
\caption{The Average MCC by LR in the Preference Elicitation Phase for CulinaryDB Dataset}
\label{fig:culinarydb-preference-elicitation}
\end{figure}
%%%%%%%%%%%%%%%%%%%%%%%%%%%%%%%%%%%%%%%%%%%%%%%%%%%%%%%%%%%%%%%%%%%%%

%%%%%%%%%%%%%%%%%%%%%%%%%%%%%%%%%%%%%%%%%%%%%%%%%%%%%%%%%%%%%%%%%%%%%
\begin{figure}[h]
\centering
\includegraphics[height=0.4\textwidth]{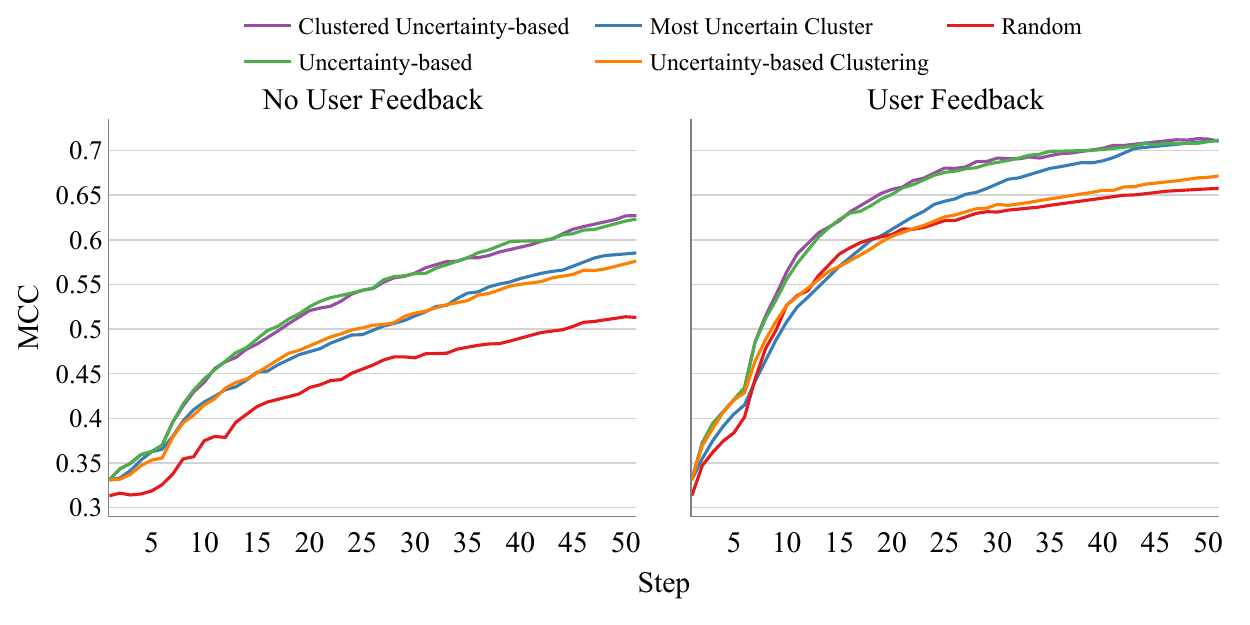}
\caption{The Average MCC by LR in 50 Interaction Steps for CulinaryDB Dataset}
\label{fig:culinarydb-50-steps}
\end{figure}
%%%%%%%%%%%%%%%%%%%%%%%%%%%%%%%%%%%%%%%%%%%%%%%%%%%%%%%%%%%%%%%%%%%%%

%%%%%%%%%%%%%%%%%%%%%%%%%%%%%%%%%%%%%%%%%%%%%%%%%%%%%%%%%%%%%%%%%%%%%
\begin{figure}[h]
\centering
\includegraphics[height=0.4\textwidth]{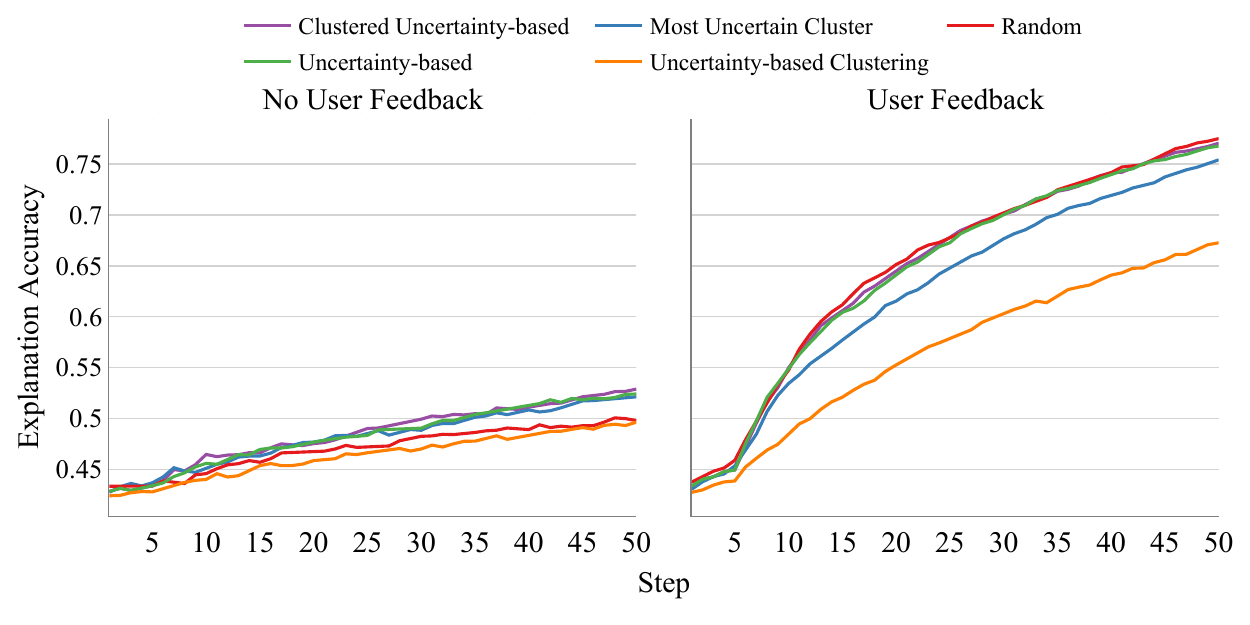}
\caption{The Explanation Accuracy of LR for CulinaryDB Dataset}
\label{fig:culinarydb-explanation-acc}
\end{figure}
%%%%%%%%%%%%%%%%%%%%%%%%%%%%%%%%%%%%%%%%%%%%%%%%%%%%%%%%%%%%%%%%%%%%%

\end{appendices}

\end{document}